\documentclass[acmlarge]{acmart}
\usepackage{multirow}
\usepackage{enumitem}
\usepackage{pifont}
\usepackage{makecell}
\usepackage[table]{xcolor}

\usepackage{graphicx}
\usepackage{subcaption}

\usepackage{booktabs} % For \toprule, \midrule, \bottomrule
\usepackage{tabularx} % For the tabularx environment and X column
\usepackage{array}    % For the >{...} syntax in \newcolumntype
\usepackage{seqsplit} % For the \seqsplit command
\usepackage{listings}

\newcommand{\bench}{\textsc{FinDeepResearch}}
\newcommand{\framework}{\emph{HisRubric}}

\newcommand{\figref}[1]{Figure \ref{#1}}
\newcommand{\tabref}[1]{Table \ref{#1}}

\newcommand{\ie}{\emph{i.e., }}
\newcommand{\eg}{\emph{e.g., }}

\newcommand{\headernodot}[1]{\noindent\textbf{#1}}
\newcommand{\header}[1]{\headernodot{$\bullet$ #1.}}

\usepackage{subcaption}

\begin{document}

%%
%% The "title" command has an optional parameter,
%% allowing the author to define a "short title" to be used in page headers.
\title{\bench: Evaluating Deep Research Agents in Rigorous Financial Analysis}

\subtitle{\textbf{Project Website}: \textcolor{blue}{\textbf{\url{https://OpenFinArena.com/}}}}

\author{Fengbin Zhu$^{*\clubsuit}$, Xiang Yao Ng$^{\spadesuit}$, Ziyang Liu$^{\spadesuit}$, Chang Liu$^{\heartsuit}$, Xianwei Zeng$^{\clubsuit}$, Chao Wang$^{\spadesuit}$,  Tianhui Tan$^{\heartsuit}$, Xuan Yao$^{\heartsuit}$, Pengyang Shao$^{\clubsuit}$,  Min Xu$^{\spadesuit}$, Zixuan Wang$^{\spadesuit}$, Jing Wang$^{\spadesuit}$,  Xin Lin$^{\spadesuit}$,  Junfeng Li$^{\clubsuit}$, Jingxian Zhu$^{\diamondsuit}$, Yang Zhang$^{\clubsuit}$, Wenjie Wang$^{\star}$, Fuli Feng$^{\star}$, Richang Hong$^{\diamondsuit}$, Huanbo Luan$^{\spadesuit}$, Ke-Wei Huang$^{\heartsuit}$, Tat-Seng Chua$^{\clubsuit}$}
\affiliation{
\institution{\\$^{\clubsuit}$National University of Singapore, Singapore\\
$^{\spadesuit}$6Estates Pte Ltd, Singapore\\
$^{\heartsuit}$Asian Institute of Digital Finance, Singapore\\
$^{\diamondsuit}$Hefei University of Technology, China\\
$^{\star}$University of Science and Technology of China}
\country{China}
}

% \author{Fengbin Zhu$^{1}$, Xiang Yao Ng$^{2}$, Ziyang Liu$^{2}$, Chang Liu$^{3}$, Xianwei Zeng$^{1}$,  Chao Wang$^{2}$,  Tianhui Tan$^{3}$, Xuan Yao$^{3}$, Pengyang Shao$^{1}$,  Min Xu$^{2}$, Zixuan Wang$^{2}$, Jing Wang$^{2}$,  Xin Lin$^{2}$,  Junfeng Li$^{1}$, Jingxian Zhu$^{4}$, Yang Zhang$^{1}$, Wenjie Wang$^{5}$, Fuli Feng$^{5}$, Richang Hong$^{4}$, Huanbo Luan$^{2}$, Ke-Wei Huang$^{3}$, Tat-Seng Chua$^{1}$}
% \affiliation{
% \institution{\\$^1$National University of Singapore,
% $^2$6Estates Pte Ltd,
% $^3$Asian Institute of Digital Finance,\\
% $^4$Hefei University of Technology,
% $^5$University of Science and Technology of China}
% }

% \author{
% Fengbin Zhu$^{\star}$, Xiang Yao Ng$^{\ast}$, Ziyang Liu$^{\ast}$, Chang Liu$^{\dagger}$, Xianwei Zeng$^{\star}$,  Chao Wang$^{\ast}$,  Tianhui Tan$^{\dagger}$, Xuan Yao$^{\dagger}$, Pengyang Shao$^{\star}$,  Min Xu$^{\ast}$, Zixuan Wang$^{\ast}$, Jing Wang$^{\ast}$,  Xin Lin$^{\ast}$,  Junfeng Li$^{\star}$, Jingxian Zhu$^{\ddagger}$, Yang Zhang$^{\star}$, Wenjie Wang$^{\circ}$, Fuli Feng$^{\circ}$, Richang Hong$^{\ddagger}$, Huanbo Luan$^{\ast}$, Ke-Wei Huang$^{\dagger}$, Tat-Seng Chua$^{\star}$}

% \affiliation{
% \institution{
% $^{\star}$National University of Singapore,
% $^{\ast}$6Estates Pte Ltd,
% $^{\dagger}$Asian Institute of Digital Finance,
% $^{\ddagger}$Hefei University of Technology,
% $^{\circ}$University of Science and Technology of China
% }
% }

\thanks{$^*$\textbf{Project Owner \& Corresponding Author: Fengbin Zhu, fengbin@nus.edu.sg}}
%%
%% The "author" command and its associated commands are used to define
%% the authors and their affiliations.
%% Of note is the shared affiliation of the first two authors, and the
%% "authornote" and "authornotemark" commands
%% used to denote shared contribution to the research.

%%
%% By default, the full list of authors will be used in the page
%% headers. Often, this list is too long, and will overlap
%% other information printed in the page headers. This command allows
%% the author to define a more concise list
%% of authors' names for this purpose.
\renewcommand{\shortauthors}{Fengbin Zhu, et al}

%%
%% The abstract is a short summary of the work to be presented in the
%% article.
\begin{abstract}
 Deep Research (DR) agents, driven by Large Language Models (LLMs), have recently garnered increasing attention for their capability in conducting complex research tasks.
 However, existing literature lacks a rigorous and systematic evaluation of DR agent's ability in critical analysis tasks. 
 To fill this gap, we first propose \textbf{\framework}, a novel evaluation framework with an expert-designed hierarchical analytical structure and a fine-grained grading rubric for rigorously assessing DR agents in corporate financial analysis.
 This framework mirrors the professional analyst's workflow, progressing from data recognition to metric calculation, and finally to strategic summarization and interpretation.
 Built on this framework, we construct a \textbf{\bench}~benchmark that comprises 64 listed companies from 8 financial markets across 4 languages, encompassing a total of 15,808 grading items.
  We further conduct extensive experiments on the \bench~ with 16 representative methods, including 6 DR agents, 5 LLMs equipped with both deep reasoning and search capabilities, and 5 LLMs with deep reasoning capabilities only.  The results reveal the strengths and limitations of these methods across diverse capabilities, financial markets, and languages, offering valuable insights for future advancements.
  The benchmark and leaderboard are publicly available on the \href{https://OpenFinArena.com/}{OpenFinArena Platform}. 
\end{abstract}

%%
%% The code below is generated by the tool at http://dl.acm.org/ccs.cfm.
%% Please copy and paste the code instead of the example below.
%%

%%
%% Keywords. The author(s) should pick words that accurately describe
%% the work being presented. Separate the keywords with commas.
% \keywords{Deep Research Agent, Deep Search, AI for Finance, Benchmarks}
%% A "teaser" image appears between the author and affiliation
%% information and the body of the document, and typically spans the
%% page.

%%
%% This command processes the author and affiliation and title
%% information and builds the first part of the formatted document.
\maketitle

\begin{figure*}[!t]  % !t 常用于让图浮到页顶
  \centering
   \setlength{\belowcaptionskip}{-0.3cm}
    \includegraphics[width=0.99\textwidth]{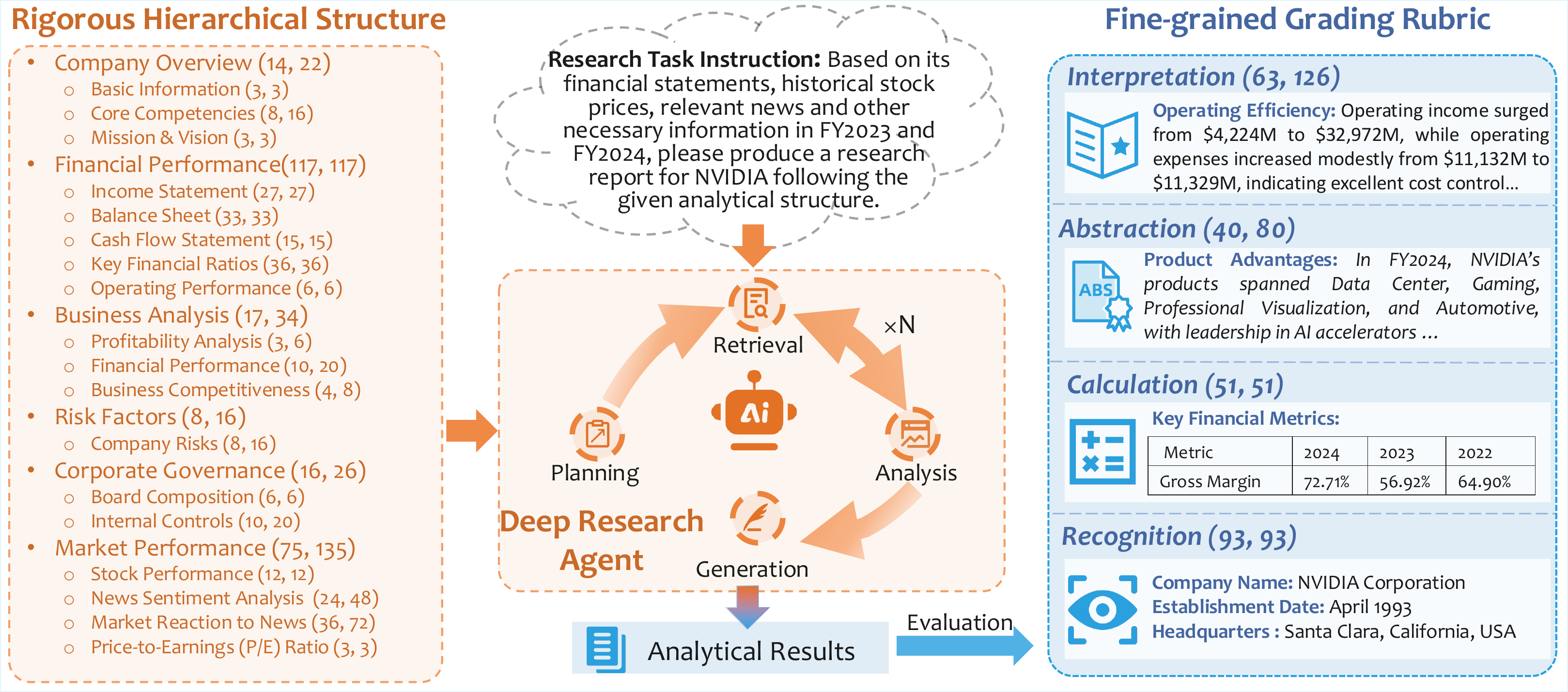}
  \caption{An overview of the \framework~evaluation framework.The numbers in brackets indicate the number of grading items (left) and the corresponding full marks (right).}
  \label{fig:framework}
\end{figure*}

\section{Introduction}
The advent of Deep Research (DR) agents, powered by the advancements in Large Language Models (LLMs), marks a pivotal shift in the ways of complex research tasks being tackled~\cite{du2025deepresearch}. 
They are capable of automatically navigating the Web, aggregating and synthesizing relevant information, and producing comprehensive reports in response to complex research tasks, such as scientific discovery~\cite{tang2025ai,zhou2025scholarsearch,liu2025researchbench} and financial analysis~\cite{sun2025finresearchbench,hu2025finsearchcomp}.
Due to such amazing capabilities, the DR agents have rapidly achieved widespread adoption~\cite{zhang2025deepresearchsurveyautonomous}, such as Gemini DR~\cite{geminiDR}, OpenAI DR~\cite{OpenAIDR}, and Grok DR~\cite{grokdeepresearch}.
Yet, rigorous and systematic approaches to evaluating their capabilities remain scarce in current literature, hindering a comprehensive understanding of their strengths and limitations.

Existing evaluation methods generally fall into two groups. 
On one hand, a line of work focuses on answer-centric verification in a Question Answering (QA) setting, reducing evaluation to a single correctness check while ignoring the substantive analysis outcomes~\cite{zhou2025scholarsearch,wei2025browsecomp,hu2025finsearchcomp,yoran2024assistantbench}. 
On the other hand, some research pursues holistic quality assessment through high-level, subjective metrics like "helpfulness" ~\cite{du2025deepresearch,sun2025finresearchbench}, or based on indeterminate report structures ~\cite{ruan2025expertlongbench,liu2025researchbench}, which yields scores that are often superficial or irreproducible.
% . This approach captures a sense of overall quality but
Thus, a fundamental tension emerges, as focusing on verifiable facts often overlooks the evaluation of analytical coherence, while an emphasis on holistic quality frequently lacks sufficient grounding in verifiable detail.
Critically, a high-quality analysis depends on both a systematic analytical structure (rigor) and specific, accurate claims (precision) simultaneously.

To overcome this, we explore a unified framework that integrates the principal goals of both research streams, which we define as two measurable criteria:
\textit{Structural Rigor}, which examines whether the agent's findings and reasoning are organized into a coherent, verifiable analytical structure; and \textit{Information Precision}, which inspects whether its claims are specific, accurate, and traceable to their sources. 
By combining these two, the framework offers a more complete and faithful measure of an agent's ability to ensure rigorous analytical quality, thereby enhancing the applicability of DR agents in critical real-world scenarios.
In this work, we ground the initial development of this framework in financial analysis, specifically focusing on corporate financial analysis. 
This task serves as an ideal testbed due to its exceptionally clear and strict professional standards. 
First, it requires a concrete analytical flow for \textit{Structural Rigor}, as analyses must follow standardized structures that cover company fundamentals, financial tables, and stock price trends, etc.
Furthermore, it provides stringent validation of  \textit{Information Precision}, demanding error-free reporting of granular details, such as year-over-year revenue growth and specific stock prices.

In this work, we introduce  \textbf{\framework}, a novel framework built on two key mechanisms: an expert-designed \textbf{\underline{Hi}}erarchical \textbf{\underline{s}}tructure to guide DR agents to conduct rigorous financial analysis and a fine-grained grading \textbf{\underline{Rubric}} for a comprehensive assessment.
%The Hierarchical Structure is designed to explicitly measure Structural Rigor, while the Rubric systematically evaluates Information Precision. 
Developed with senior financial experts, our hierarchical structure defines a practical analytical structure for corporate financial analysis, comprising $6$ major sections and $18$ subsections.
The Rubric is composed of $247$ fine-grained grading items designed to assess 4 progressive capabilities of DR agents, \ie \emph{Recognition}, \emph{Calculation}, \emph{Abstraction}, and \emph{Interpretation}. These dimensions align closely with established evaluative frameworks for financial analysis and the quality of analyst reports from an academic perspective \cite{herath2017financial,gaynor2016understanding}, and are also consistent with best practices recognized in global financial markets from an industry perspective. In practice, leading institutions such as \emph{Institutional Investor}'s \emph{All-America Research Team Awards} and \emph{Refinitiv StarMine Analyst Awards} systematically evaluate research quality based on cognitive accuracy, reasoning depth, and interpretive insight, while the \emph{CFA Institute's Graham \& Dodd Awards} highlight excellence in applied financial analysis and communication clarity.\footnote{See Institutional Investor Research Awards: \url{https://www.institutionalinvestor.com/research}; Refinitiv StarMine Awards: \url{https://www.refinitiv.com/en/star-mine}; CFA Institute Graham \& Dodd Awards: \url{https://rpc.cfainstitute.org/research/financial-analysts-journal/graham-and-dodd-awards-of-excellence}.} Together, these industry standards reinforce the relevance of the four dimensions as key indicators of analytical rigor and professional competence.

With the \framework~framework, we construct a \textbf{\bench}~benchmark, encompassing companies from $8$ financial markets (\ie United States, United Kingdom, China, Hong Kong, Australia, Singapore, Malaysia, and Indonesia ) across $4$ languages (\ie English, Simplified Chinese, Traditional Chinese, Bahasa Indonesia). 
From each financial market, we select $8$ companies, resulting in a total of $64$ listed companies with $15,808$ grading items.
These companies are distributed across $10$ industries, defined by the Bloomberg Industry Classification Standard (BICS), including  Communications, Energy,  Health Care, Materials, and Technology, etc. 
For each company, DR agents are required to generate a research report that follows the hierarchical analytical structure and is grounded in a diverse set of data, such as financial statements, stock prices, financial news, market indices, etc.

% exp 
On the constructed \bench~benchmark, we conduct extensive experiments with 16 representative methods, including 6 DR agents, 5 LLMs with thinking and search capabilities, and 5 LLMs with thinking capability only.
The experimental results reveal that: 
1) Most methods generally conform to the expert-designed analytical structure, but they consistently fall short in generating precise information.
2) DR agents consistently exhibit superior performance  compared to the methods in the other two categories, with their advantage being particularly pronounced in the Recognition and Calculation capabilities.
3) All evaluated methods face significant challenges in mastering the Interpretation capabilities and in performing corporate financial analysis of non-English markets.

In summary, the major contributions of this work are threefold:
\begin{itemize}[leftmargin=*, nosep]
    \item We introduce a novel \framework~evaluation framework built upon a practical hierarchical analytical structure and a fine-grained grading rubric for assessing Deep Research agents in critical and rigorous financial analysis. 
    \item We construct a \bench~benchmark comprising companies from $8$ financial markets across $4$ languages, resulting in a total of $64$ listed companies with $15,808$ grading items.
    \item We conduct extensive experiments on \bench~with 16 models, including advanced DR agents and representative LLMs equipped with web search and/or deep reasoning capabilities. The results indicate that while most methods successfully adhere to the prescribed analytical structure, they consistently struggle with producing precise information.
\end{itemize}

\section{\framework~Framework}
In \figref{fig:framework}, we present the proposed \framework~evaluation framework, which integrates an expert-defined hierarchical analytical structure with a fine-grained grading rubric to systematically assess recognition, calculation, abstraction, and interpretation capabilities of deep research methods.

\subsection{Task Definition}

% \subsubsection{Task Description} 
To ensure the high standard of the research outcomes, we devise a comprehensive hierarchical analytical structure to guide the analysis. 
Formally, given a research task instruction $i$ with a desired analytical structure $S$, a method $\mathcal{M}$ is required to produce a research report $\mathcal{R}$ strictly following the analytical structure $S$. 
\begin{equation}    
  \setlength{\abovecaptionskip}{-0.4pt}
\setlength{\belowcaptionskip}{-0.3cm}
    \label{eq:task}
    \mathcal{R} = \mathcal{M}(i, S)
\end{equation}.

In this work, the instruction $i$ is provided in natural language.
Both the analytical structure $S$ and the generated research report $R$ are formatted in Markdown to facilitate easy evaluation.

\subsection{Rigorous Hierarchical Structure}

\begin{figure*}[!t]  % !t 常用于让图浮到页顶
  \centering
  \setlength{\abovecaptionskip}{-0.4pt}
\setlength{\belowcaptionskip}{-0.3cm}
  \includegraphics[width=0.96\textwidth]{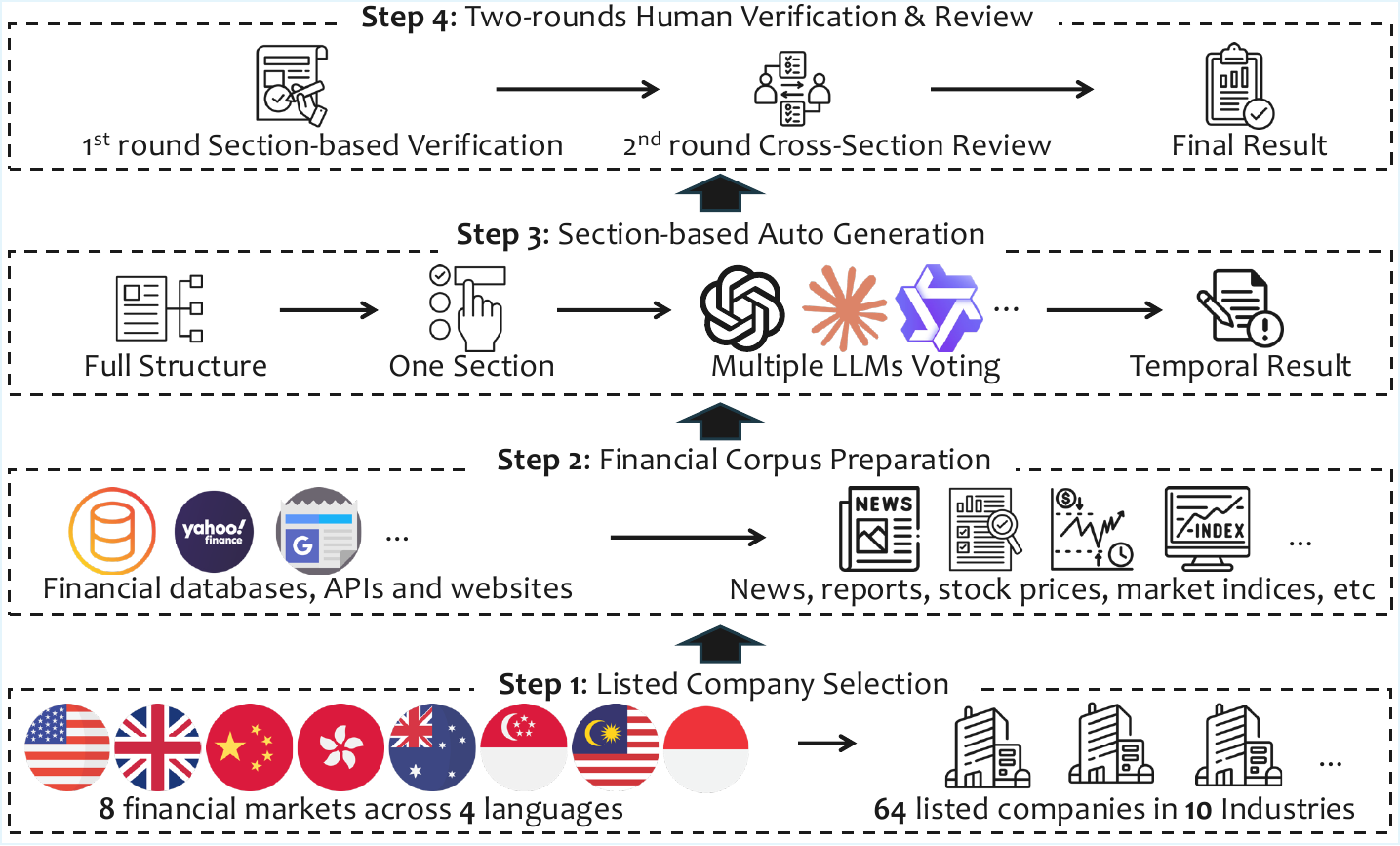}
  \caption{An overview for constructing \bench.}
  \label{fig:bench-overview}
  \vspace{-0.5cm}
\end{figure*}

As shown in Figure~\ref{fig:framework}, to achieve a comprehensive and rigorous evaluation, we employ proficient financial experts to devise a practical
hierarchical analytical structure for corporate finance analysis with 6 major sections and 18 subsections, covering the key perspectives in real-world corporate analysis as follows: 

\begin{itemize}[leftmargin=*, nosep]
    \item  \textbf{Section 1: Company Overview.} This section provides a concise overview of the company, including its basic information, industry background, key strengths, and strategic direction.
    It is divided into $3$ subsections: Basic Information, Core Competencies, and Mission \& Vision.
    
    \item  \textbf{Section 2: Financial Performance.} This section presents a detailed analysis of the company's financial health, including primary financial statements and key performance metrics.
    It comprises $5$ subsections: Income Statement, Balance Sheet, Cash Flow Statement, Key Financial Ratios, and Operating Performance.

\item  \textbf{Section 3: Business Analysis.} Through a deep analysis of the obtained data, this section identifies key insights regarding the company's business, financial performance, and profitability.
This section includes $3$ subsections: Profitability Analysis, Financial Performance Summary, and Business Competitiveness.

\item  \textbf{Section 4: Risk Factors.} This section identifies and discusses the principal risks the company faces, including market, financial, operational, and regulatory risks, along with the strategies in place to manage them.

\item  \textbf{Section 5: Corporate Governance.} This section outlines the company's governance framework, including the board of directors, executive leadership, governance policies, and practices, ensuring transparency and accountability. 
This section contains $2$ subsections: Board Composition and Internal Controls. 

\item  \textbf{Section 6: Market Performance.} This section provides a comprehensive analysis of the company's stock performance, the news events that shape its public narrative, and its current market valuation. 
It is structured into $4$ subsections: Stock Performance, News Sentiment Analysis, Market Reaction to News, and Price-to-Earnings Ratio.

\end{itemize}

\subsection{Fine-grained Grading Rubric}

To facilitate a comprehensive evaluation of the generated financial research report, a fine-grained grading rubric is applied.
From each section in the structure, we select specific data items for scoring, termed ``\emph{grading items}'', which are designed to ensure full coverage of all key analytical perspectives. 
Each of these grading items is then mapped to one of four critical capabilities of DR agents:

\begin{itemize}[leftmargin=*, nosep]
    \item  \textbf{Recognition.} The capability to accurately identify and extract specific factual data from vast and complex data sources, serving as a fundamental skill.
    
\item  \textbf{Calculation.} The ability to precisely compute and verify numerical values, which is essential for rigorous quantitative analysis.

\item  \textbf{Abstraction.} One critical competency to synthesize complex relationships and summarize valuable patterns, enabling the distillation of essential perspectives from messy data.

\item  \textbf{Interpretation.} The capacity to conduct deep analysis on the existing data to deliver insightful findings and implications, reflecting the highest level of reasoning.
\end{itemize}

In total, we obtain $247$ distinct grading items, and the distribution across the four capabilities is presented on the right of \figref{fig:framework}.
According to financial expert assessment, the competencies assessed under Abstraction and Interpretation are more complex than those under the other two categories. Consequently, items in Abstraction and Interpretation are weighted at 2 marks each, while items in the remaining categories are weighted at 1 mark each. This weighting scheme yields a total possible score of 350 marks.
 
\begin{table}[t]
\centering
\caption{Statistics of \bench.}
\setlength{\tabcolsep}{15mm}
% \footnotesize
\begin{tabular}{lr}
\toprule
\textbf{Statistic} & \textbf{Number} \\
\hline
\multicolumn{2}{l}{\textcolor{black}{\textbf{\textit{Basic Information}}}} \\
% \color{red}{} \\
\: Number of Languages & 4 \\
\: Number of Financial Markets & 8 \\
\: Number of Industries & 10 \\
\: Number of Selected Companies & 64 \\
% \hline
\multicolumn{2}{l}{\textcolor{black}{\textbf{\textit{Analytical Structure}}}} \\
\: Number of Major Sections  & 6 \\
\: Number of Subsections & 18 \\
% \hline
\multicolumn{2}{l}{\textcolor{black}{\textbf{\textit{Grading Items}}}} \\
\: Number of Grading Items per Report & 247 \\
\: Full Marks for each Report & 350 \\
\: Total Number of Grading Items & 15,808 \\
\bottomrule
\end{tabular}%
\label{tab:dataset_statistic}
\vspace{-0.55cm}
\end{table}

\subsection{Evaluation Protocol}

To assess the \emph{Information Precision}, for each grading item, we first obtain the predicted answer from the generated result and then compare it with the corresponding ground truth.
Three distinct evaluation protocols are applied to different types of grading items. 
\begin{itemize}[leftmargin=*,nosep]
    \item \textit{Accuracy}: We employ an advanced LLM to evaluate the correctness by comparing the predicted answer to the ground truth. It gives a score of 1 for a match, otherwise 0. This method is applied to all grading items in \emph{Recognition} and \emph{Calculation} and to a subset of items with concrete answers in \emph{Interpretation}.
    \item  \textit{Claim-based Score}: We first adopt an advanced LLM to identify three to five critical reference claims from the ground truth, depending on the length of the ground truth. For each claim, we apply the LLM to determine whether it is adequately covered in the predicted answer\cite{Ip_deepeval_2025}. The proportion of covered claims constitutes this claim-based score, ranging from 0 to 1. This method is applied to all grading items in \emph{Abstraction} and to a subset of items formed in a summary format in \emph{Interpretation}.
    \item \textit{Criterion-based Score}: For items requiring nuanced reasoning and qualitative analysis, a simple binary or claim-based evaluation is insufficient. We therefore introduce a criterion-based scoring approach\cite{zhang2023evaluating}.
    This process begins by prompting an advanced LLM to act as the role of a financial expert (\eg a financial professor) to generate a detailed 10-point scoring criterion based on the ground truth. This criterion deconstructs the ideal answer into its core analytical components. Subsequently, the LLM is used to grade the predicted answer against the criterion. The final score is the sum of the awarded points, normalized to a scale of 0 to 1.
    This method is applied to a subset of the \emph{Interpretation} items where the quality of argumentation and the depth of analysis are key assessment factors.
\end{itemize}
After summing the scores from all grading items, the total score is normalized by the maximum possible value (\ie 350) to yield a final score ranging from 0 to 1, termed ``accuracy score''.

In addition, we also assess the \emph{Structural Rigor} of the generated markdown result with a rule-based validation method.
Our method evaluates structural compliance by scoring the 6 main sections, 18 subsections, and 18 markdown tables. The scoring awards 1 point for each correct element and deducts 1 point for errors, yielding a format score out of a maximum of 42 points. 
The raw score is then normalized by the maximum (\ie 42) to produce a final score between 0 and 1, termed ``structure score'', which provides a quantitative measure of structural fidelity.

\begin{table}[t]
\centering
\caption{Analysis of \bench~ across 8 financial markets.}
\setlength{\abovecaptionskip}{-0.6pt}
\setlength{\belowcaptionskip}{-0.8cm}
% \setlength{\tabcolsep}{0.3mm}
% \footnotesize
\begin{tabular}{lrrrrrrrr}
\toprule
\multirow{2}{*}{\bf Metric} & \includegraphics[width=0.5cm]{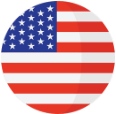} &\includegraphics[width=0.5cm]{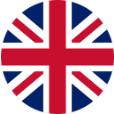} & \includegraphics[width=0.5cm]{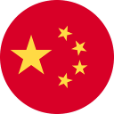} & \includegraphics[width=0.5cm]{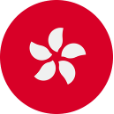} & \includegraphics[width=0.5cm]{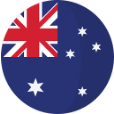} & \includegraphics[width=0.5cm]{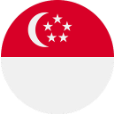} & \includegraphics[width=0.5cm]{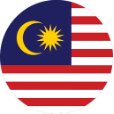} &
\includegraphics[width=0.5cm]{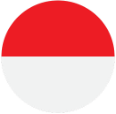}  \\
  & \bf US & \bf UK & \bf CN & \bf HK & \bf AU & \bf SG & \bf MY & \bf ID \\
\hline
\#Selected Companies & 8 & 8& 8& 8& 8& 8& 8& 8 \\ 
\\
Min \#Chars per Company & 31,985 & 21,605 & 15,040 & 11,940 & 20,358 & 16,034 & 18,683 & 27,535 \\
Avg \#Chars per Company & 46,032 & 31,029 & 22,928 & 26,680 & 25,218 & 26,420 & 24,320 & 30,528 \\
Med \#Chars per Company & 40,017 & 30,544 & 20,835 & 23,625 & 23,123 & 23,560 & 21,784 & 29,455 \\
Max \#Chars per Company & 82,257 & 40,041 & 42,713 & 46,967 & 36,169 & 45,036 & 37,740 & 36,181 \\
\\
Min \#Chars per Grading Item & 3 & 3 & 3 & 2 & 2 & 3 & 3 & 2 \\
Avg \#Chars per Grading Item & 173 & 112 & 78 & 93 & 88 & 94 & 85 & 109 \\
Med \#Chars per Grading Item & 17 & 16 & 20 & 18 & 16 & 18 & 19 & 18 \\
Max \#Chars per Grading Item & 3,803 & 3,526 & 2,590 & 3,159 & 2,332 & 3,940 & 1,898 & 2,235 \\

\bottomrule
\end{tabular}
\label{tab:dataset_distribution}
\vspace{-0.1cm}
\end{table}

\section{\bench~Benchmark}
This section introduces the construction and quality control of \bench, presents a statistical analysis of its properties, and compares it with existing deep research benchmarks.

\subsection{Construction of \bench}
We illustrate the overall pipeline for constructing our \bench~in \figref{fig:bench-overview}, including four key steps:

\begin{itemize}[leftmargin=*, nosep]
\item \textbf{Step 1: Listed Company Selection}
To ensure a comprehensive and diverse evaluation, we select companies from eight financial markets: the United States (US), the United Kingdom (UK), China (CN), Hong Kong (HK), Australia (AU), Singapore (SG), Malaysia (MY), and Indonesia (ID). 
This selection covers four languages: English (EN), Simplified Chinese (zh-CN), Traditional Chinese (zh-HK), and Bahasa Indonesia (BI).
Finally, we obtain 64 listed companies from 10 industries (\ie Property \& Real Estate, Healthcare \& Communications, Consumer Discretionary, Consumer Staples, Energy, Health Care, Industrials, Materials, Real Estate, Technology, Utilities) based on the Bloomberg Industry Classification Standard
(BICS).

\item \textbf{Step 2: Financial Corpus Preparation}
After the companies are selected, we obtain the associated financial data from a variety of data providers, including established financial databases (\eg Bloomberg), API services (\eg Alpha Vantage), Google News), and public financial websites (\eg Yahoo Finance~\footnote{Bloomberg:\url{https://www.bloomberg.com/}; Alpha Vantage:\url{https://www.alphavantage.co/}; Google News: \url{https://gnews.io/}; Yahoo Finance:\url{https://finance.yahoo.com/}}. 
The collected data includes fundamental data, annual reports, historical stock prices and market indices, and relevant news, etc.

\item \textbf{Step 3: Section-based Auto Generation}
Next, we generate a reference report for each selected company. 
For every section in the expert-designed structure, multiple Large Language Models (LLMs) are leveraged to take the relevant corpus as input and generate candidate results separately.
The predominant result for each grading item in the section is then selected among the multiple candidates as the definitive value.
Upon completion of all sections, these results are synthesized into a provisional full report for subsequent verification.

\item \textbf{Step 4: Two-rounds Human Verification \& Review}
Finally, our financial experts conduct two rounds of data verification.
To enhance consistency and efficiency of the human review process, we conduct a section-based verification technique in the first round. 
First, we divide the financial experts into $6$ groups, with each group responsible for verifying a specific section.
This round is concluded only after all sections have been verified.
In the second round, a panel of senior financial experts is assigned to review the entire report, performing a cross-verification of all sections to ensure both consistency and accuracy.

\end{itemize}

\begin{table}[t]
\centering
\caption{Comparison between our \bench~and other deep research benchmarks.}
\small
\begin{tabular}{llcrrc}
\toprule
\bf Name & \bf Domain & \bf Structured &  \bf Languages & \bf \#Answers/Items & \bf Retrieval Corpus \\
\hline
GAIA~\cite{mialon2023gaia} & General & \ding{53} & EN & 466 & Web \\
BrowseComp~\cite{wei2025browsecomp} & General & \ding{53} & EN & 1,266 & Web \\
AssistantBench~\cite{yoran2024assistantbench} & General & \ding{53} & EN & 214 & Web \\
ExpertLongBench~\cite{ruan2025expertlongbench} & General & \ding{53} & EN & Various & Offline \\
DeepResearchBench~\cite{du2025deepresearch} & General & \ding{53} & EN,zh-CN & 2,500 & Web \\
ScholarSearch~\cite{zhou2025scholarsearch} & Academic & \ding{53} & EN & 223 & Offline \\
ResearchBench~\cite{liu2025researchbench} & Academic & \ding{53} & EN & 678 & Web \\
FinSearchComp~\cite{hu2025finsearchcomp} & Finance & \ding{53} & EN, zh-CN & 635 & Web \\
FinResearchBench~\cite{sun2025finresearchbench} & Finance & \ding{53} & EN & Various & Web \\
\hline
\bf \bench & Finance & \ding{52} & EN, zh-CN, zh-HK, BI & 15,808 & Web \\
\bottomrule
\end{tabular}
\label{tab:comparison}
\vspace{-0.2cm}
\end{table}

\subsection{Quality Control}
We maintain the high quality of \bench~by implementing a rigorous quality-control process throughout its construction, including,

\header{Proficient Financial Experts} 
The cohort of financial experts comprises over $30$ professional practitioners, academic researchers, and graduate students in economics and related disciplines from leading institutions and universities.
These experts are deeply involved in the entire benchmark construction process, from structure design to section verification and report review. 
To ensure structural rigor and practical applicability, a dedicated senior team comprising industry experts with over ten years of experience, finance professors, and postdoctoral researchers, is assembled to design the analytical structure.
Subsequent to the generation of the reference report, the details of each report undergo verification by financial experts. 

\header{Cross-source Data Validation} 
The acquisition of financial data for the benchmark construction is drawn from a multi-source framework, including proprietary financial databases, official corporate websites, and established financial portals with API services. To ensure the integrity and accuracy of critical data, such as financial tables, key financial ratios, stock prices, and market indices, a cross-source validation protocol is implemented. Under this protocol, an individual data point is incorporated into the dataset only if it is corroborated by a minimum of two independent sources.
In instances where discrepancies arise, a manual review is conducted by financial experts to arbitrate and determine the final value. This cross-source validation approach mitigates the risk of systematic errors and inconsistencies, thereby safeguarding the high quality of the benchmark.

\header{Rigorous Structure Guided} The construction of the benchmark is guided by a comprehensive and rigorous analytical structure designed by financial experts. 
The clearly defined and unambiguous grading items within this structure facilitate high-quality result generation and systematic verification.

\header{Two-rounds Expert Verification} To ensure the high quality of the benchmark, financial experts conduct a two-round verification process that includes both section-based error correction and report-based consistency checks. This approach guarantees that each grading item is reviewed by minimum two financial experts.

\begin{figure}[!t]  % !t 常用于让图浮到页顶
  \centering
  \includegraphics[width=0.9\textwidth]{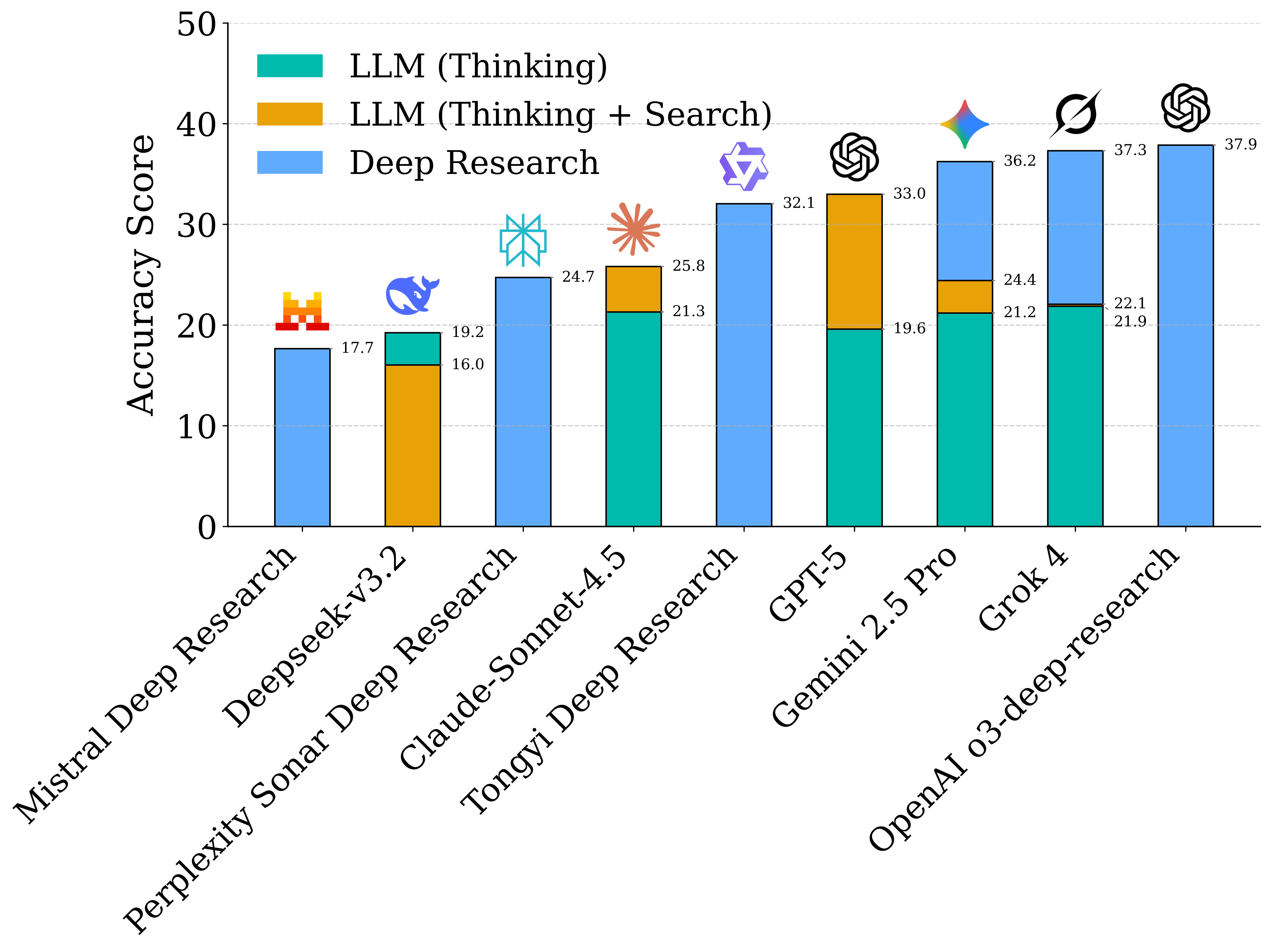}
  \caption{An evaluation of representative methods on \bench~ w.r.t  \emph{Information Precision}.}
  \label{fig:main-result}
  % \vspace{-0.3cm}
\end{figure}

\subsection{Statistic and Analysis}

As shown in \tabref{tab:dataset_statistic}, the \bench~dataset comprises 64 companies spanning 10 industries across 8 financial markets in 4 languages.
Each company's analysis is structured hierarchically into 6 major sections, further subdivided into 18 subsections. 
For quantitative assessment, 247 data points are selected from each analysis and incorporated into a scoring system totaling 350 marks.
Consequently, the entire benchmark encompasses 15,808 individual grading items.

To ensure a fair evaluation, we select 8 companies from each financial market.
For each market, \tabref{tab:dataset_distribution} summarizes the character count statistics for the reference analytical report of each company and for the answers of each grading item.

\subsection{Comparison with Other Benchmarks}

Table \ref{tab:comparison} presents a comparative analysis of \bench~against existing deep research benchmarks, highlighting its key advantages. Our benchmark differentiates itself in three key aspects: 1) Whereas existing benchmarks do not require analytically structured outputs, \bench~mandates that responses adhere to a rigorous and predefined structure. 2) It offers superior multilingual coverage; while most related works are limited to English, and others like DeepResearchBench~\cite{du2025deepresearch} and FinSearchComp~\cite{hu2025finsearchcomp} include only English and Simplified Chinese, our dataset encompasses four languages: English, Simplified Chinese, Traditional Chinese (Hong Kong), and Bahasa Indonesia. 3) With 15,808 data items for scoring, \bench~significantly surpasses the scale of prior benchmarks, enabling a more comprehensive and robust evaluation.

\begin{figure}[!t]  % !t 常用于让图浮到页顶
  \centering
  \includegraphics[width=0.98\textwidth]{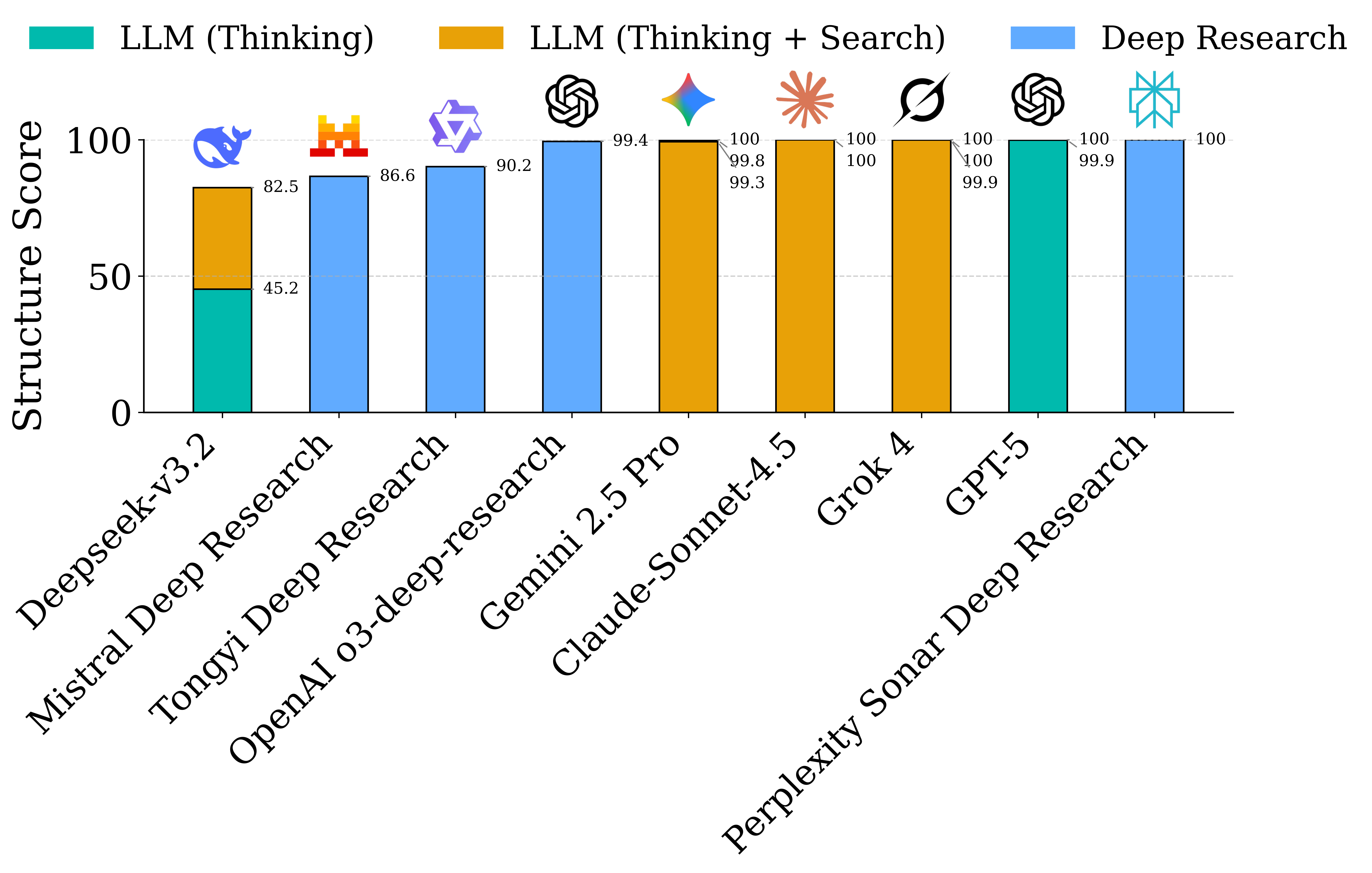}
  \caption{An evaluation of representative methods on \bench~w.r.t \emph{Structural Rigor}.}
  \label{fig:structure}
  % \vspace{-0.3cm}
\end{figure}

\section{Experiments}
In this section, we introduce the experimental setup and present the comprehensive analysis of the experimental results.

\begin{figure*}[t]
  \centering
  \setlength{\belowcaptionskip}{-0.05cm}
  \begin{subfigure}[b]{0.46\textwidth}
    \centering
    
  \setlength{\abovecaptionskip}{-0.1pt}
  \setlength{\belowcaptionskip}{-0.1cm}
    \includegraphics[width=\linewidth]{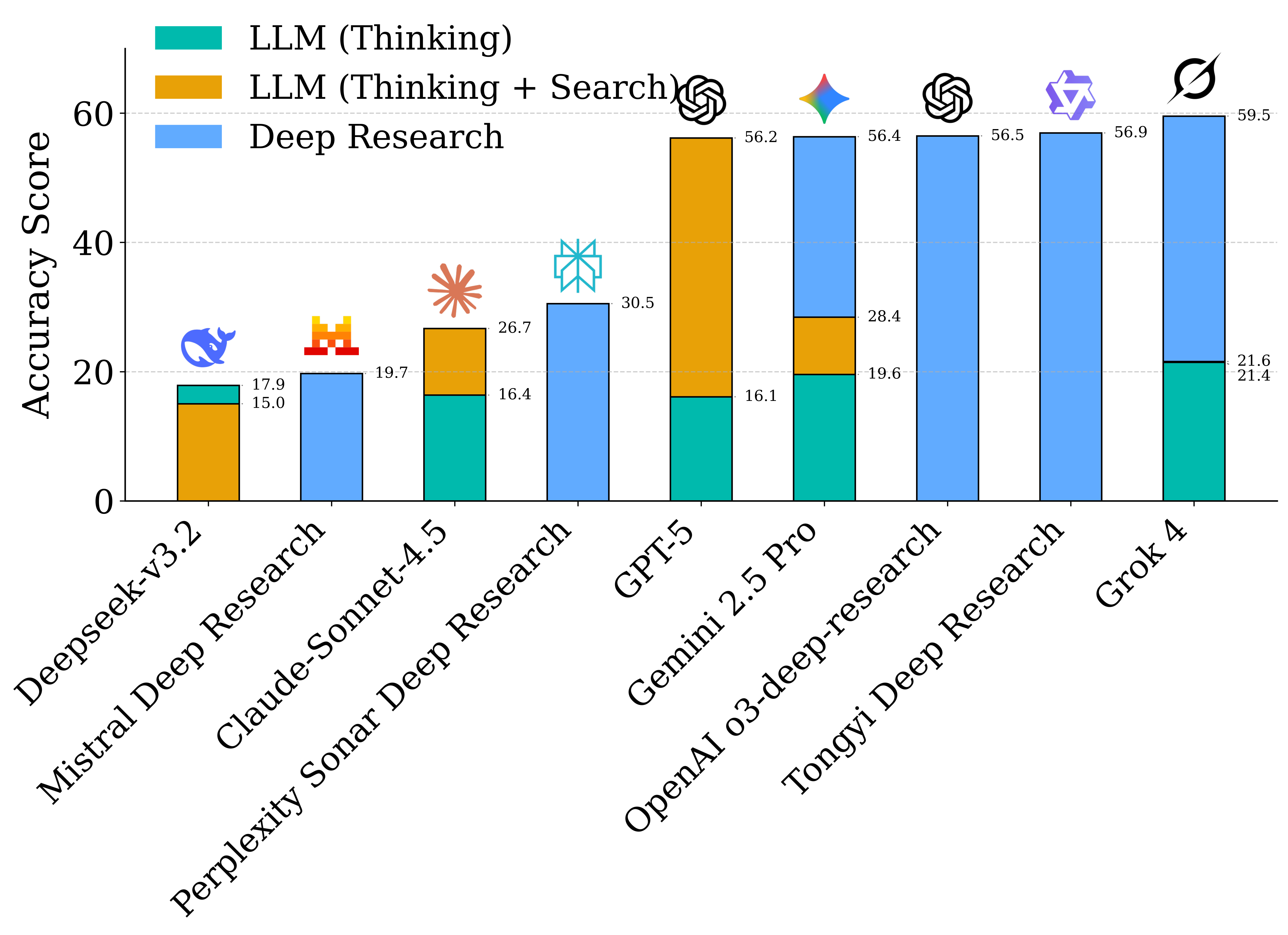}
    \caption{Recognition}
    \label{fig:recognition}
  \end{subfigure}
  \hfill
  \begin{subfigure}[b]{0.46\textwidth}
    \centering
     \setlength{\abovecaptionskip}{-0.6pt}
  \setlength{\belowcaptionskip}{-0.1cm}
    \includegraphics[width=\linewidth]{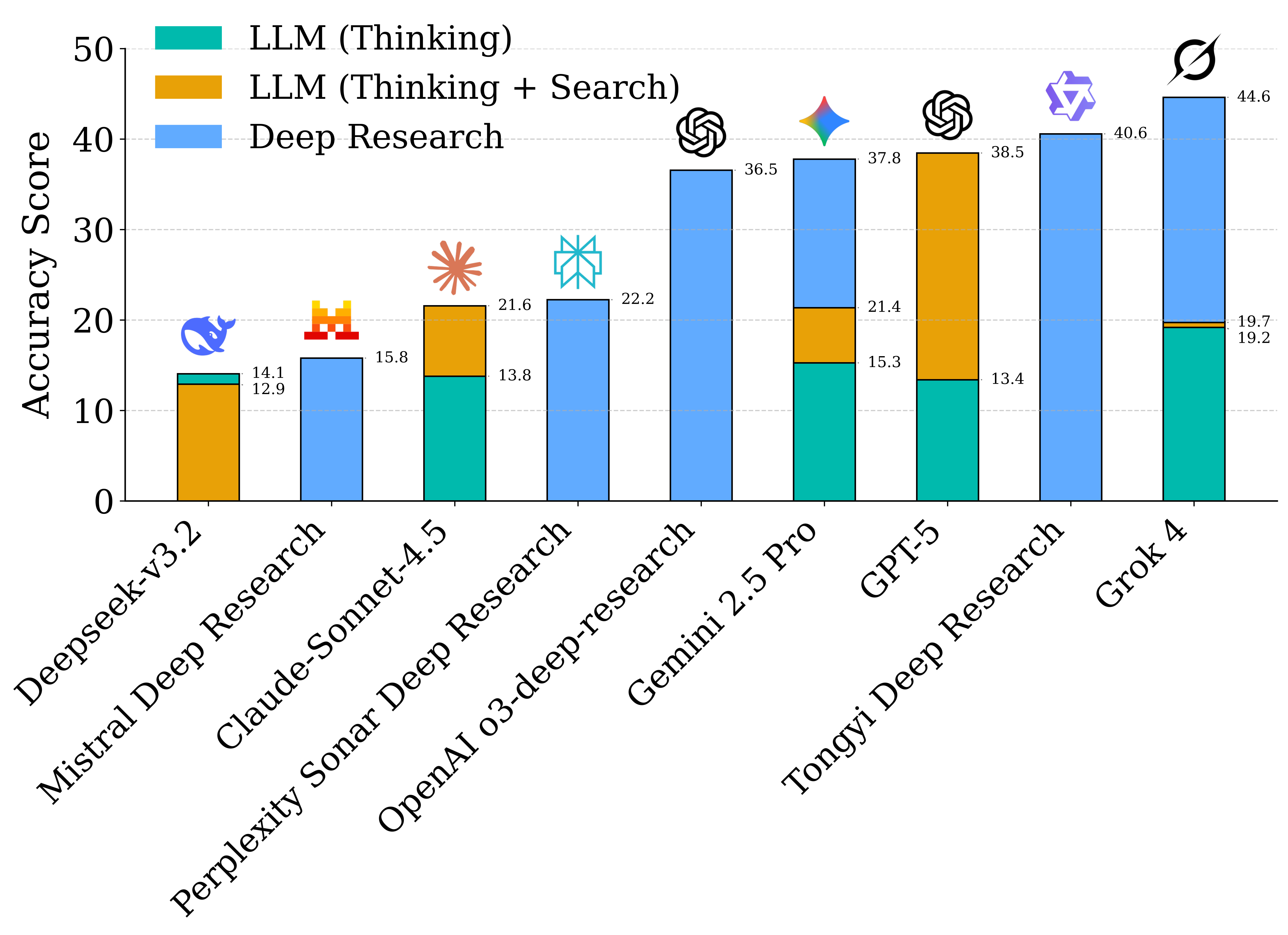}
    \caption{Calculation}
    \label{fig:calculation}
  \end{subfigure}
  \begin{subfigure}[b]{0.46\textwidth}
    \centering
     \setlength{\abovecaptionskip}{-0.6pt}
  \setlength{\belowcaptionskip}{-0.1cm}
    \includegraphics[width=\linewidth]{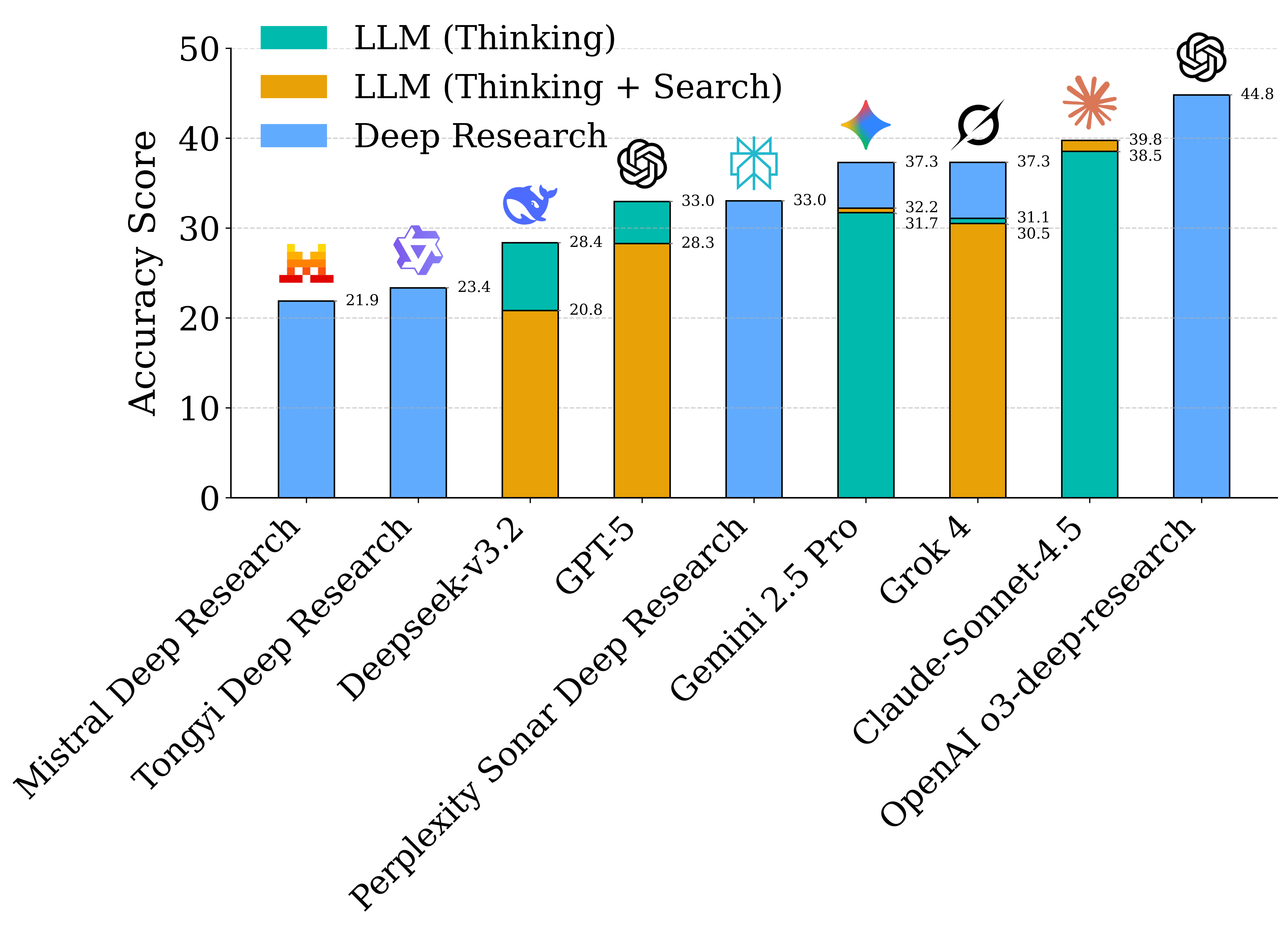}
    \caption{Abstraction}
    \label{fig:abstraction}
  \end{subfigure}
  \hfill
  \begin{subfigure}[b]{0.46\textwidth}
    \centering
     \setlength{\abovecaptionskip}{-0.6pt}
      \setlength{\belowcaptionskip}{-0.1cm}
    \includegraphics[width=\linewidth]{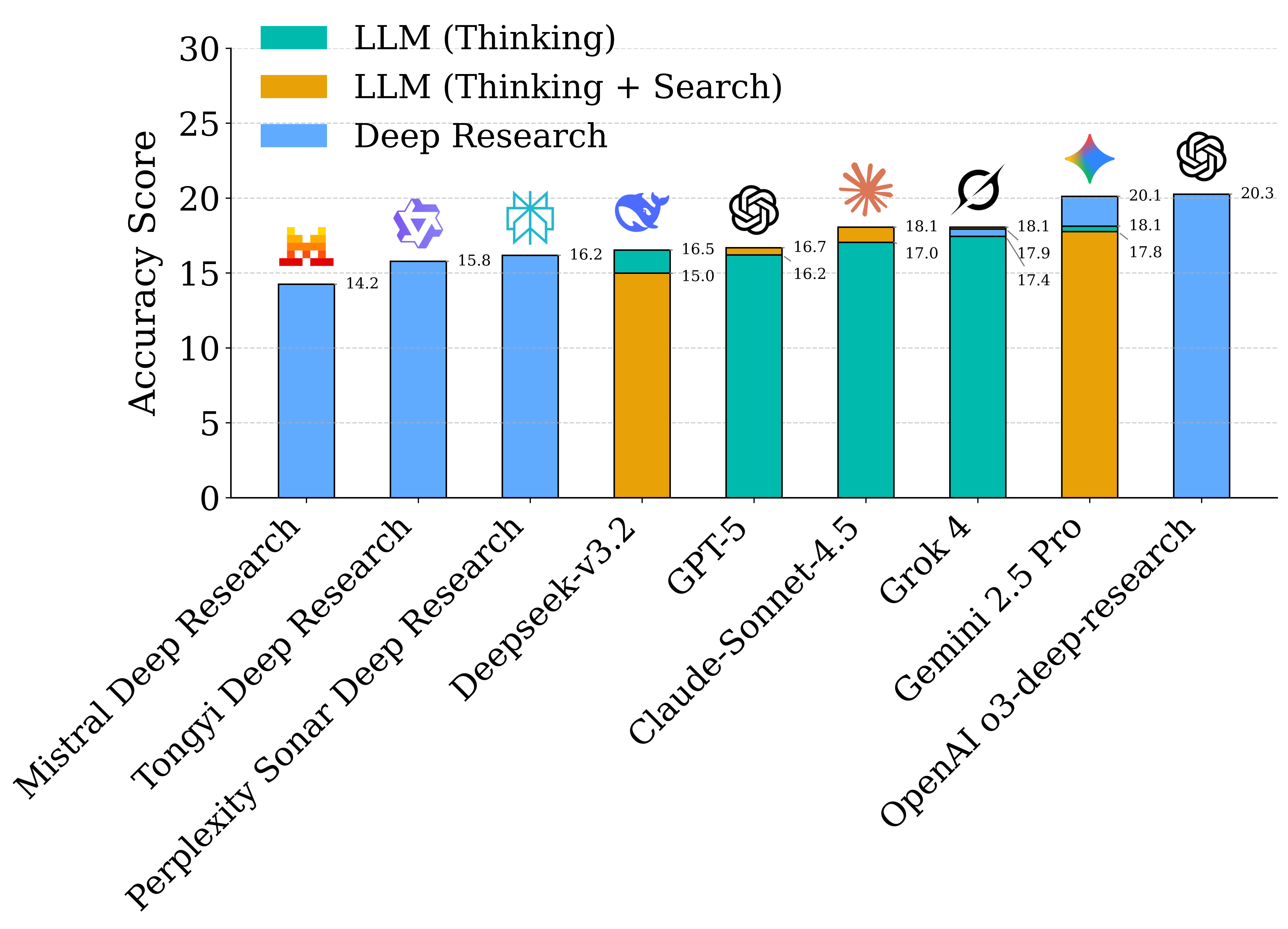}
    \caption{Interpretation}
    \label{fig:interpretation}
  \end{subfigure}
  \caption{Performance analysis across four different capabilities.}
  \label{fig:four-tasks}
  \vspace{-0.2cm}
\end{figure*}

\subsection{Compared Methods} With the proposed \bench, we conduct experiments with 16 methods from 3 different groups.
The selected methods exhibit considerable diversity in their underlying model families.

\header{LLM with Thinking (T)}
OpenAI GPT-5 (T)~\cite{gpt5}, Claude-Sonnet-4.5 (T)~\cite{claudesonnet45}, Gemini 2.5 Pro (T)~\cite{gemini25pro}, Deepseek-v3.2 (T)~\cite{deepseekv32}, and Grok 4 (T)~\cite{grok4DR}.

\header{LLM with Thinking + Search (T+S)} 
OpenAI GPT-5 (T+S)~\cite{gpt5}, Claude-Sonnet-4.5 (T+S)~\cite{claudesonnet45}, Gemini 2.5 Pro (T+S)~~\cite{gemini25pro}, Deepseek-v3.2 (T+S)~~\cite{deepseekv32}, and Grok 4 (T+S)~~\cite{grok4DR}.

\header{Deep Research} OpenAI o3-deep-research~\cite{OpenAIDR}, Gemini 2.5 Pro Deep Research~\cite{geminiDR}, Grok 4 DeepSearch~\cite{grok4DR}, 
Perplexity Sonar Deep Research~\cite{perplexitydeepresearch},
Tongyi Deep Research~\cite{tongyiDR} and Mistral Deep Research~\cite{mistralDR}.

\subsection{Main Results}
We present a comparable analysis of different methods on our proposed \bench, assessing their performance in terms of \emph{Information Precision} and \emph{Structural Rigor}.

\header{Information Precision}
\figref{fig:main-result} illustrates a comparison of all methods with respect to \emph{Information Precision}. 
We make the following key findings: 
1) Among all methods, OpenAI's o3-deep-research achieves the highest performance with an accuracy score of 37.9. 
It is closely followed by Grok-4 DeepSearch, which ranks a competitive second with a score of 37.3.
2) Deep Research (DR) methods generally perform better than the other two types. 
This is clearly seen in the top five results, where four are DR methods, and the remaining slot is held by OpenAI GPT-5 (T+S), their most advanced LLM.
This result demonstrates the superiority of DR methods in solving high-standard and complex analysis tasks like our proposed \bench.
3) LLMs relying solely on deep reasoning perform poorly, which underscores the necessity of search capability to retrieve external knowledge for effectively addressing the challenges in \bench.
4) All methods face significant challenges in addressing \bench, as evidenced by the highest score achieved being only 37.9 out of a maximum of 100, clearly indicating the persistent difficulty of the benchmark.

\header{Structural Rigor} \figref{fig:structure} shows the performance of all methods with respect to \emph{Structural Rigor}.
We make the following observations:
1) Most of the methods can produce the analytical results strictly following the predefined hierarchical structure.
Of all evaluated methods, 7 of them generate outputs with perfect formatting, and 5 of them contain only minor formatting errors.
The findings suggest that advanced LLMs have developed the capability to follow complex instructions, such as the hierarchical structure in this study, which is fundamental for successfully executing rigorous research tasks. 
2) From \figref{fig:main-result} and \figref{fig:structure}, we can observe that methods that perform poorly in structure following generally exhibit suboptimal results in generating accurate information.
For instance, DeepSeek-v3.2 (T), DeepSeek-v3.2 (T+S), and Mistral Deep Research rank as the bottom three in \emph{Structural Rigor}, and also show the weakest performance in \emph{Information Precision}.

\begin{table}[t]
\centering
\caption{The performance analysis across 8 financial markets. The values reported in the table denote the normalized accuracy score. The best and second-best methods are indicated with \textbf{bold} and \underline{underline}, respectively.}
\setlength{\tabcolsep}{3.5mm}
% \small
\begin{tabular}{lrrrrrrrr}
\toprule
\multirow{2}{*}{\bf Method} & \includegraphics[width=0.5cm]{icons/US.png} &\includegraphics[width=0.5cm]{icons/UK.png} & \includegraphics[width=0.5cm]{icons/CN.png} & \includegraphics[width=0.5cm]{icons/HK.png} & \includegraphics[width=0.5cm]{icons/AU.png} & \includegraphics[width=0.5cm]{icons/SG.png} & \includegraphics[width=0.5cm]{icons/MY.png} &
\includegraphics[width=0.5cm]{icons/ID.png}  \\
  & \bf US & \bf UK & \bf CN & \bf HK & \bf AU & \bf SG & \bf MY & \bf ID \\
\hline
\rowcolor{lightgray} \multicolumn{9}{c}{\textcolor{black}{\emph{\textbf{LLM (Thinking)}}}} \\

Gemini 2.5 Pro (T) & 19.9 & 21.0 & 17.6 & 20.8 & 24.4 & 24.2 & 25.1 & 16.5 \\
Deepseek-v3.2 (T) & 19.7 & 17.7 & 17.3 & 18.4 & 20.9 & 21.0 & 23.8 & 15.0 \\
Claude-Sonnet-4.5 (T) & 22.2 & 19.9 & 19.1 & 21.7 & 23.0 & 22.7 & 24.7 & 17.0 \\
Grok 4 (T) & 23.2 & 24.0 & 16.9 & 18.4 & 25.8 & 24.3 & 25.0 & 17.4 \\
OpenAI GPT-5 (T) & 18.1 & 18.7 & 16.6 & 17.6 & 22.6 & 23.6 & 23.3 & 16.3 \\

\rowcolor{lightgray}\multicolumn{9}{c}{\textcolor{black}{\emph{\textbf{LLM (Thinking + Search)}}}} \\
Gemini 2.5 Pro (T+S) & 22.9 & 20.7 & 20.4 & 24.7 & 26.4 & 27.6 & 27.5 & 20.9 \\
Deepseek-v3.2 (T+S) & 10.9 & 14.9 & 16.8 & 16.5 & 20.4 & 17.7 & 21.0 & 10.0 \\
Claude-Sonnet-4.5 (T+S) & 27.8 & 23.0 & 25.7 & 20.3 & 27.4 & 28.5 & 30.4 & 23.4 \\
Grok 4 (T+S) & 23.7 & 22.4 & 17.8 & 19.4 & 27.2 & 24.6 & 25.0 & 16.4 \\
OpenAI GPT-5 (T+S) & 37.4 & 36.9 & 20.8 & 29.3 & 35.6 & \underline{42.5} & 32.3 & 29.1 \\

\rowcolor{lightgray}\multicolumn{9}{c}{\textcolor{black}{\emph{\textbf{Deep Research}}}} \\
Perplexity Sonar Deep Research & 21.0 & 23.7 & 22.4 & 25.0 & 28.8 & 26.9 & 26.9 & 23.0 \\
Mistral Deep Research & 13.5 & 16.1 & 14.0 & 13.6 & 22.2 & 21.1 & 23.7 & 17.1 \\
Tongyi Deep Research & 32.1 & 27.8 & 27.8 & 29.5 & 36.1 & 35.6 & 37.3 & 30.3 \\
Gemini 2.5 Pro Deep Research & \underline{37.6} & 34.1 & 30.8 & \underline{36.0} & 36.0 & 38.9 & \bf 39.8 & \underline{36.6} \\
Grok 4 DeepSearch & 34.5 & \underline{39.0} & \underline{33.4} & \bf 36.4 & \underline{39.3} & \bf 46.7 & 37.9 & 31.3 \\
OpenAI o3-deep-research & \bf 42.5 & \bf 43.0 & \bf 34.7 & 30.2 & \bf 41.7 & 33.6 & \underline{38.3} & \bf 38.9 \\

\bottomrule
\end{tabular}
\label{tab:markets}
\vspace{-0.7cm}
\end{table}

\subsection{In-depth Analysis}

\header{Performance Analysis Across Markets}
\tabref{tab:markets} reports the model performance in accuracy score across 8 financial markets and reveals the following findings: 
1) Deep Research methods exhibit superior performance across all eight evaluated markets, significantly surpassing the results of ``Thinking'' and ``Thinking + Search'' approaches. 
For instance, OpenAI o3-deep-research leads five markets (\ie US, UK, CN, AU, ID), whereas Grok 4 DeepSearch secures the top rank in both HK and SG, and Gemini 2.5 Pro Deep Research achieves the highest performance in the MY market.
Notably, Open-sourced Tongyi Deep Research demonstrates competitive performance against proprietary DR methods and outperforms most ``Thinking'' and ``Thinking + Search'' approaches. 
These findings collectively affirm the superiority of DR methods for addressing such high-standard and complex research analysis.
2) A significant performance gap exists across markets. 
Specifically, CN and HK present a tougher challenge, evidenced by their peak scores of only 34.7 by OpenAI o3-deep-research and 36.4 by Grok 4 DeepSearch, compared to easier markets like SG (46.7), UK (43.0), and US (42.5). 
This difficulty may stem from the increased complexity that methods encounter when processing languages other than the Latin languages, including Simplified Chinese and Traditional Chinese.
3) None of the markets are fully addressed by existing methods. 
Even in the best-performing SG market, Grok 4 DeepSearch achieves only 46.7 in the accuracy score. This sizable distance from the perfect score of 100 indicates substantial headroom for future advances in the field.

\header{Performance Analysis Across Capabilities}

\figref{fig:four-tasks} presents a comparative performance analysis of the models across the four critical capabilities: Recognition, Calculation, Abstraction, and Interpretation.
We make the following key findings: 
1) Grok 4 DeepSearch ranks first for the Recognition and Calculation capabilities, while OpenAI o3-deep-research achieves the highest performance in the Abstraction and Interpretation capabilities.
2) DR methods generally outperform the methods in the other two categories for Recognition and Calculation capabilities. 
Comparably, the performance gap narrows significantly across all models for Abstraction and Interpretation.
3) Performance across the four capabilities demonstrates a clear difficulty spectrum, with Recognition being the most effectively addressed capability, achieving the highest score of 59.5. Calculation and Abstraction exhibit comparable performance, peaking at 44.6 and 44.8, respectively. Conversely, Interpretation is currently the most difficult, evidenced by its maximum score of only 20.3.
This significant lag suggests that improving Interpretation capability should be a promising focus for future research.
4) All four capabilities remain far from a perfect score. 
The highest performance achieved, 59.5 by Grok 4 DeepSearch in Recognition, indicates that there is substantial room for overall improvement.

\begin{figure}[!t]  % !t 常用于让图浮到页顶
  \centering
  \setlength{\abovecaptionskip}{-0.6pt}
\setlength{\belowcaptionskip}{-0.5cm}
\includegraphics[width=0.98\textwidth]{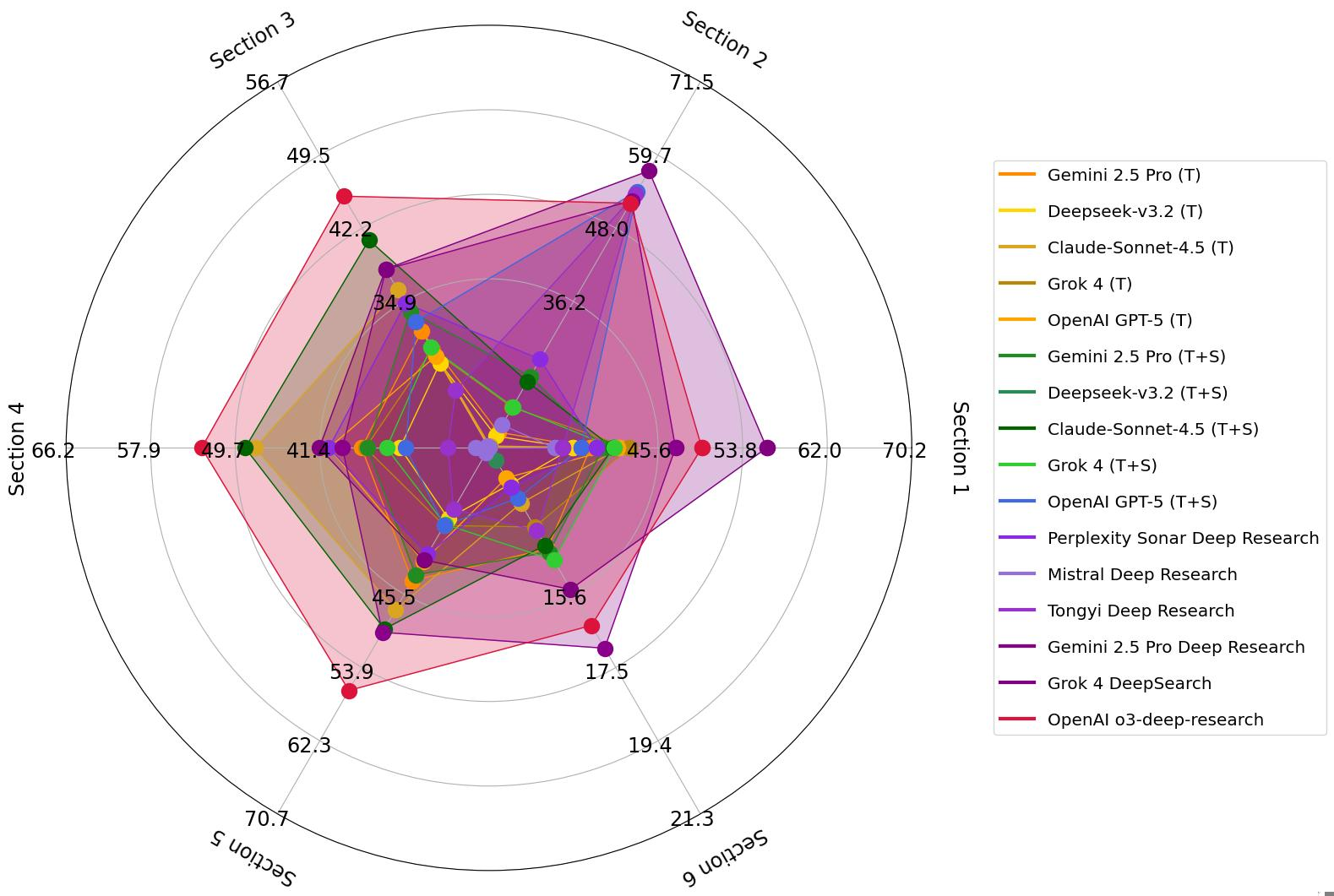}
  \caption{Performance analysis across sections. To ensure comparability across different sections, the values reported represent the normalized accuracy score, capped at 100.}
  \vspace{-0.1cm}
  \label{fig:section-perf}
\end{figure}

\header{Performance Analysis Across Sections}
\figref{fig:section-perf} shows the performance analysis across different sections. 
We make the following observations: 
1) Performance varies significantly across the sections. Sections 1, 2, 4, and 5 show moderate success with the best normalized accuracy score above 50, and Section 3 reaches 45.4, while Section 6 presents a severe challenge, with a maximum accuracy score of only 17.0. This low performance of Section 6 is attributed to the demands for integrated analysis of diverse inputs, including financial statements, stock prices, news, market indices, and currency exchange rates.
2) Deep Research methods exhibit clear advantages. Specifically, Grok-4 Deep Research achieves the highest scores in Sections 1 and 2, OpenAI o3 Deep Research attains the best performance in Sections 3, 4, and 5, while Gemini 2.5 Pro Deep Research leads in Section 6. This demonstrates the superior efficacy of the Deep Research methods in significantly enhancing report quality compared to the other two methodologies.
3) Across all sections, LLM (Thinking + Search) methods generally demonstrate superior performance over LLM (Thinking) methods, highlighting the importance of retrieval for this task. This superiority is most evident in sections 2 and 3.

\subsection{Case Study}
We evaluate the three approaches, \ie "Thinking (T)" , "Thinking + Search (T+S)", and "Deep Research (DR)", to characterize their performance boundaries. In Table~\ref{tab:case_study}, we use tick marks (\ding{51}) to denote good cases where metrics are generally accurately retrieved or calculated and cross marks (\ding{55}) to indicate bad cases where results are mostly inaccurate or unavailable. Results are organized into three approaches based on performance patterns: 
\begin{itemize}[leftmargin=*, nosep]
    \item \textbf{T (\ding{55}), T+S(\ding{51}), DR (\ding{51})}: T produces bad results while T+S and DR produce good results. This performance gap arises because the T approach operates solely on the model's parametric knowledge without access to external information sources, limiting its ability to accurately extract financial data from relevant data sources and subsequently compute metrics. In contrast, both T+S and DR methods leverage external retrieval capabilities to access requisite financial data.
    \item \textbf{T (\ding{55}), T+S(\ding{55}), DR (\ding{51})}: Both T and T+S produce bad results while DR produces good results. The superior performance of DR can be attributed to its agent-driven architecture, which facilitates iterative retrieval and reasoning capabilities. The iterative nature of DR allows for self-correction and refinement through multiple reasoning cycles, leading to higher quality outputs compared to the single-pass reasoning employed by T and T+S approaches.
    \item \textbf{T (\ding{55}), T+S(\ding{55}), DR (\ding{55})}: All three approaches yield unsatisfactory results. These shortcomings arise primarily in metrics that require historical adjusted closing price data over extended horizons (typically one year), which are difficult to obtain with precision. A further source of error stems from the demands for integrated analysis of diverse data sources, including financial statements, stock prices, news, market indices, and currency exchange rates.

\end{itemize}

\begin{table}[t]
\centering
\caption{Case Study. ``T'', ``T+S'' and ``DR'' represent ``LLM (Thinking)'', ``LLM (Thinking + Search)'' and ``Deep Research''.}
\label{tab:case_study}
\scriptsize
% \begin{tabular}{|p{3cm}|p{5cm}|p{7cm}|}
\begin{tabular}{l|l|p{10cm}}
\toprule
\textbf{Type} & \textbf{Grading Items} & \textbf{Example} \\
\hline
\multirow{4}{*}{ \makecell[l]{T (\ding{55})\\ T+S(\ding{51})\\ DR (\ding{51}) \\}}  & \multirow{4}{*}{ \makecell[l]{Income before Income Taxes  \\ Total Liabilities \\ Shareholders Equity \\...}} & Income before Income Taxes: \textcolor{blue}{1,721 Millions USD} \\
& & OpenAI GPT-5 (T): \textcolor{red}{N/A} \\
& & OpenAI GPT-5 (T+S): \textcolor{blue}{1,721 in millions USD} \\
& & OpenAI o3-deep-research: \textcolor{blue}{1,721 Millions USD} \\
 
\hline
\multirow{10}{*}{ \makecell[l]{T (\ding{55})\\ T+S(\ding{55})\\ DR (\ding{51}) \\}}  & \multirow{10}{*}{ \makecell[l]{Profitability and Earnings Quality \\ Market Position \\ Core Values \\ ...}} & Profitability and Earnings Quality: \textcolor{blue}{Profitability declined in 2024, with a decrease in net profit margin and gross margin. The net profit also decreased significantly.} \\
& & OpenAI GPT-5 (T): \textcolor{red}{- - Profitability compressed due to raw-material inflation but remained ... earnings quality supported by cash conversion and low reliance on non-recurring items... margins trailed the prior year’s levels.} \\
& & OpenAI GPT-5 (T+S): \textcolor{red}{Margin compression ... Earnings quality supported by positive operating cash flow (US\$52.6m) despite headwinds.
} \\
& & OpenAI o3-deep-research: \textcolor{blue}{Profitability slipped in 2024, with net income down ~26\% . Earnings quality remained decent (all profit derived from core operations), but margins were squeezed. Gross profit decline and higher expenses ...} \\

\hline
\multirow{4}{*}{ \makecell[l]{T (\ding{55})\\ T+S(\ding{55})\\ DR (\ding{55})}}  & \multirow{4}{*}{ \makecell[l]{Annualized Volatility \\ Log Excess Return \\ Maximum Drawdown \\...}} & Annualized Volatility: \textcolor{blue}    {17.40\%} \\
& & OpenAI GPT-5 (T): \textcolor{red}{N/A} \\
& & OpenAI GPT-5 (T+S): \textcolor{red}{N/A} \\
& & OpenAI o3-deep-research: \textcolor{red}{20.10\% }\\
\bottomrule
\end{tabular}
\end{table}

\section{Related Work}

\subsection{Deep Research Agents}

LLMs have recently shown strong capabilities in reasoning and problem solving, motivating the development of DR agents that autonomously explore the web and generate research reports~\cite{guo2025deepseek,qwq32b,zeng2025glm}. Among early agents, ReAct~\cite{yao2023react} is among the earliest to couple reasoning traces with environment actions, enabling interleaved reasoning-and-acting for open-ended tasks. Building on this idea, Search-R1 uses reinforcement learning to decide when and how to issue search queries for multi-hop question answering~\cite{jin2025searchr}, and MMSearch-R1 extends this line by incorporating multimodal search for joint text–image reasoning~\cite{wu2025mmsearch}. While effective, these methods generally do not produce well-structured and comprehensive research reports.

To address this gap, recent DR agents integrate planning, multi-round retrieval, and evidence-grounded synthesis in a dynamic loop~\cite{kimideepresearch,zhang2025web,grokdeepresearch,perplexitydeepresearch}. For example, the Gemini 2.5 Pro Deep Research agent~\cite{geminiDR} plans research, performs broad-coverage retrievals, and synthesizes a structured report end-to-end after reinforcement-learning-driven fine-tuning. OpenAI Deep Research ~\cite{OpenAIDR} provides a ChatGPT-based workflow that interactively clarifies queries, browses the live web, analyzes retrieved content with built-in tools, and produces source-grounded, citation-rich summaries. Qwen Deep Research~\cite{qwendeepresearch} employs dynamic research blueprinting and concurrent task orchestration to improve autonomous planning and adaptive execution. 
Despite these advances, rigorously evaluating the structured research outcomes generated by DR agents remains a major challenge, as there is still no consensus on how to measure both their structural completeness and information accuracy.

\subsection{Benchmarks for Deep Research Agents}

Benchmarking DR agents has become a critical avenue for assessing their ability to plan, retrieve, and synthesize evidence into structured research reports~\cite{wan2025deepresearcharena,yoran2024assistantbench}. General-purpose benchmarks typically evaluate agents on open-domain problems requiring long-horizon reasoning, factual grounding, and iterative synthesis~\cite{liu2025researchbench,wan2025deepresearcharena,wei2025browsecomp,ruan2025expertlongbench}. Among them, DeepResearch Bench~\cite{du2025deepresearch} spans diverse academic disciplines and employs structured frameworks such as RACE and FACT to measure report comprehensiveness, instruction following, and citation fidelity; ExpertLongBench~\cite{ruan2025expertlongbench} targets expert-level long-form outputs through checklist-based rubrics.

Domain-specific benchmarks, in contrast, focus on emphasizing professional expertise, time sensitivity, and finer-grained evaluation~\cite{hu2025finsearchcomp,sun2025finresearchbench}.
For instance, FinSearchComp~\cite{hu2025finsearchcomp} emphasizes financial analyst workflows: retrieving real-time market data, performing historical lookups, and conducting multi-period investigations with expert-annotated tasks and a rigorous multi-stage QA process. FinResearchBench~\cite{sun2025finresearchbench}, evaluates financial research agents by extracting logic trees from their reports and assessing performance across 70 expert-curated questions spanning 7 key task types. Yet even these domain-specific efforts remain limited: most confine evaluation to short-form answers or coarse global report scores and rarely assess full-length research reports in critical analysis scenarios for structural completeness, evidence reconciliation, and fine-grained factual accuracy.
To the best of our knowledge, this work is the first to propose \framework, a novel evaluation framework and \bench~benchmark for rigorously assessing deep research agents in financial analysis.

\section{Conclusion}

In this paper, we introduce \framework, a novel framework to evaluate the ability of DR agents to conduct high-quality and rigorous financial analysis, by defining and measuring the core qualities of \textit{Structural Rigor} and \textit{Information Precision}. We construct a new benchmark,  \bench, covering 64 companies across 8 markets and 4 languages. Our experiments suggest that even top-performing DR agents struggle to consistently balance a coherent analytical structure with factual accuracy. This imbalance remains the primary barrier to their deployment in high-stakes applications.
Future work can extend our framework to other domains, such as legal and clinical research, and explore how novel agent architectures might narrow this performance gap. In summary, we contend that a dual evaluation of rigor and precision is a crucial step towards building the next generation of reliable DR agents for professional, real-world tasks.

\section{Contributions}

\begin{itemize}
    \item \textbf{Project Leader:} Fengbin Zhu, Chao Wang, and Tianhui Tan.
    \item  \textbf{Major Contributors:} Xiang Yao Ng, Ziyang Liu, Chang Liu, Xianwei Zeng, Xuan Yao, and Min Xu.
    \item  \textbf{Secondary Contributors:} Zixuan Wang, Pengyang Shao, Jing Wang, Xin Lin, Junfeng Li, Jingxian Zhu, and Yang Zhang.
    \item  \textbf{Advisors:} Wenjie Wang, Fuli Feng, Richang Hong, Huanbo Luan, Ke-Wei Huang, and Tat-Seng Chua.
\end{itemize}

%%
%% The acknowledgments section is defined using the "acks" environment
%% (and NOT an unnumbered section). This ensures the proper
%% identification of the section in the article metadata, and the
%% consistent spelling of the heading.
% \begin{acks}
% To Robert, for the bagels and explaining CMYK and color spaces.
% \end{acks}

%%
%% The next two lines define the bibliography style to be used, and
%% the bibliography file.
\bibliographystyle{ACM-Reference-Format}
\bibliography{sample-base}

@misc{zhang2025deepresearchsurveyautonomous,
      title={Deep Research: A Survey of Autonomous Research Agents}, 
      author={Wenlin Zhang and Xiaopeng Li and Yingyi Zhang and Pengyue Jia and Yichao Wang and Huifeng Guo and Yong Liu and Xiangyu Zhao},
      year={2025},
      eprint={2508.12752},
      archivePrefix={arXiv},
      primaryClass={cs.IR},
      url={https://arxiv.org/abs/2508.12752}, 
}

@article{herath2017financial,
  title={Financial reporting quality: A literature review},
  author={Herath, Siriyama and Albarqi, Norah},
  journal={Journal of Business Management and Commerce},
  volume={2},
  pages={1--14},
  year={2017}
}

@article{gaynor2016understanding,
  title={Understanding the relation between financial reporting quality and audit quality},
  author={Gaynor, Lisa Milici and Kelton, Andrea Seaton and Mercer, Molly and Yohn, Teri Lombardi},
  journal={Auditing: A Journal of Practice \& Theory},
  volume={35},
  number={4},
  pages={1--22},
  year={2016},
  publisher={American Accounting Association},
  doi={10.2308/ajpt-51453}
}

@article{du2025deepresearch,
  title={DeepResearch Bench: A Comprehensive Benchmark for Deep Research Agents},
  author={Du, Mingxuan and Xu, Benfeng and Zhu, Chiwei and Wang, Xiaorui and Mao, Zhendong},
  journal={arXiv preprint arXiv:2506.11763},
  year={2025}
}

@article{wan2025deepresearcharena,
  title={DeepResearch Arena: The First Exam of LLMs' Research Abilities via Seminar-Grounded Tasks},
  author={Wan, Haiyuan and Yang, Chen and Yu, Junchi and Tu, Meiqi and Lu, Jiaxuan and Yu, Di and Cao, Jianbao and Gao, Ben and Xie, Jiaqing and Wang, Aoran and others},
  journal={arXiv preprint arXiv:2509.01396},
  year={2025}
}

@misc{zhou2025scholarsearch,
      title={ScholarSearch: Benchmarking Scholar Searching Ability of LLMs}, 
      author={Junting Zhou and Wang Li and Yiyan Liao and Nengyuan Zhang and Tingjia Miao and Zhihui Qi and Yuhan Wu and Tong Yang},
      year={2025},
      eprint={2506.13784},
      archivePrefix={arXiv},
      primaryClass={cs.IR},
      url={https://arxiv.org/abs/2506.13784}, 
}

@article{wei2025browsecomp,
  title={Browsecomp: A simple yet challenging benchmark for browsing agents},
  author={Wei, Jason and Sun, Zhiqing and Papay, Spencer and McKinney, Scott and Han, Jeffrey and Fulford, Isa and Chung, Hyung Won and Passos, Alex Tachard and Fedus, William and Glaese, Amelia},
  journal={arXiv preprint arXiv:2504.12516},
  year={2025}
}

@inproceedings{yoran2024assistantbench,
  title={AssistantBench: Can Web Agents Solve Realistic and Time-Consuming Tasks?},
  author={Yoran, Ori and Amouyal, Samuel and Malaviya, Chaitanya and Bogin, Ben and Press, Ofir and Berant, Jonathan},
  booktitle={Proceedings of the 2024 Conference on Empirical Methods in Natural Language Processing},
  pages={8938--8968},
  year={2024}
}

@article{ruan2025expertlongbench,
  title={ExpertLongBench: Benchmarking Language Models on Expert-Level Long-Form Generation Tasks with Structured Checklists},
  author={Ruan, Jie and Nair, Inderjeet and Cao, Shuyang and Liu, Amy and Munir, Sheza and Pollens-Dempsey, Micah and Chiang, Tiffany and Kates, Lucy and David, Nicholas and Chen, Sihan and others},
  journal={arXiv preprint arXiv:2506.01241},
  year={2025}
}

@article{sun2025finresearchbench,
  title={FinResearchBench: A Logic Tree based Agent-as-a-Judge Evaluation Framework for Financial Research Agents},
  author={Sun, Rui and Bai, Zuo and Zhang, Wentao and Zhang, Yuxiang and Zhao, Li and Sun, Shan and Qiu, Zhengwen},
  journal={arXiv preprint arXiv:2507.16248},
  year={2025}
}

@article{liu2025researchbench,
  title={Researchbench: Benchmarking llms in scientific discovery via inspiration-based task decomposition},
  author={Liu, Yujie and Yang, Zonglin and Xie, Tong and Ni, Jinjie and Gao, Ben and Li, Yuqiang and Tang, Shixiang and Ouyang, Wanli and Cambria, Erik and Zhou, Dongzhan},
  journal={arXiv preprint arXiv:2503.21248},
  year={2025}
}

@article{hu2025finsearchcomp,
  title={FinSearchComp: Towards a Realistic, Expert-Level Evaluation of Financial Search and Reasoning},
  author={Hu, Liang and Jiao, Jianpeng and Liu, Jiashuo and Ren, Yanle and Wen, Zhoufutu and Zhang, Kaiyuan and Zhang, Xuanliang and Gao, Xiang and He, Tianci and Hu, Fei and others},
  journal={arXiv preprint arXiv:2509.13160},
  year={2025}
}

@inproceedings{mialon2023gaia,
  title={Gaia: a benchmark for general ai assistants},
  author={Mialon, Gr{\'e}goire and Fourrier, Cl{\'e}mentine and Wolf, Thomas and LeCun, Yann and Scialom, Thomas},
  booktitle={The Twelfth International Conference on Learning Representations},
  year={2023}
}

@article{geminiDR,
      title={Deep Research is now available on Gemini 2.5 Pro Experimental},
      author={Dave Citron},
      journal={https://blog.google/products/gemini/deep-research-gemini-2-5-pro-experimental/},
      year={2025}
}

@misc{grokdeepresearch,
  author = {{xAI Team}},
  title = {Introducing Grok DeepSearch},
  year = {2025},
  howpublished = {\url{https://x.ai/news/grok-3}},
  note = {Accessed: 2025-04-06}
}

@misc{OpenAIDR,
    author = {{OpenAI Team}},
    title = {Introducing deep research},
    year = {2025},
    howpublished = {\url{https://openai.com/index/introducing-deep-research/}},
    note = {Accessed: 2025-10-07}
}

@misc{grok4DR,
  author = {{xAI Team}},
  title = {Grok 4},
  year = {2025},
  howpublished = {\url{https://x.ai/news/grok-4}},
  note = {Accessed: 2025-10-07}
}

@misc{tongyiDR,
    author =  {{Tongyi Team}},
    title = {Tongyi DeepResearch: A New Era of Open-Source AI Researchers},
    year = {2025},
    howpublished = {\url{https://tongyi-agent.github.io/blog/introducing-tongyi-deep-research/}},
    note = {Accessed: 2025-10-07}
}

@misc{mistralDR,
    author =  {{Mistral Team}},
    title = {TLe Chat dives deep},
    year = {2025},
    howpublished = {\url{https://mistral.ai/news/le-chat-dives-deep}},
    note = {Accessed: 2025-10-07}
}

@misc{perplexitydeepresearch,
  author = {Perplexity Team},
  title = {Introducing perplexity deep research},
  year = {2025},
  howpublished = {\url{https://www.perplexity.ai/ja/hub/blog/introducing-perplexity-deep-research}}
}

@misc{qwendeepresearch,
  author = {Qwen Team},
  title = {Deep research (Qwen-Deep-Research)},
  year = {2025},
  howpublished = {\url{https://www.alibabacloud.com/help/en/model-studio/qwen-deep-research}}
}

@misc{kimideepresearch,
  author = {Kimi Team},
  title = {Kimi-Researcher: End-to-End RL Training for Emerging Agentic Capabilities},
  year = {2025},
  howpublished = {\url{https://moonshotai.github.io/Kimi-Researcher/?utm_source=chatgpt.com}}
}

@misc{gpt5,
    author = {OpenAI Team},
    title = {Introducing GPT-5},
    year = {2025},
    howpublished = {\url{https://openai.com/index/introducing-gpt-5/}},
    note = {Accessed: 2025-10-07}
}

@misc{gemini25pro,
    author = {Gemini Team} ,
    title = {Gemini 2.5 Pro},
    year = {2025},
    howpublished = {\url{https://deepmind.google/models/gemini/pro/}},
    note = {Accessed: 2025-10-07}
}

@misc{claudesonnet45,
    author = {Claude Team} ,
    title = {Introducing Claude Sonnet 4.5},
    year = {2025},
    howpublished = {\url{https://www.anthropic.com/news/claude-sonnet-4-5}},
    note = {Accessed: 2025-10-07}
}

@misc{deepseekv32,
    author = {DeepSeek Team} ,
    title = {Introducing DeepSeek-V3.2-Exp},
    year = {2025},
    howpublished = {\url{https://api-docs.deepseek.com/news/news250929}},
    note = {Accessed: 2025-10-07}
}

@inproceedings{yao2023react,
  title={React: Synergizing reasoning and acting in language models},
  author={Yao, Shunyu and Zhao, Jeffrey and Yu, Dian and Du, Nan and Shafran, Izhak and Narasimhan, Karthik and Cao, Yuan},
  booktitle={International Conference on Learning Representations (ICLR)},
  year={2023}
}

@inproceedings{
    jin2025searchr,
    title={Search-R1: Training {LLM}s to Reason and Leverage Search Engines with Reinforcement Learning},
    author={Bowen Jin and Hansi Zeng and Zhenrui Yue and Jinsung Yoon and Sercan O Arik and Dong Wang and Hamed Zamani and Jiawei Han},
    booktitle={Second Conference on Language Modeling},
    year={2025},
}

@article{tang2025ai,
  title={AI-Researcher: Autonomous Scientific Innovation},
  author={Tang, Jiabin and Xia, Lianghao and Li, Zhonghang and Huang, Chao},
  journal={arXiv preprint arXiv:2505.18705},
  year={2025}
}

@article{wu2025mmsearch,
  title={MMSearch-R1: Incentivizing LMMs to Search},
  author={Wu, Jinming and Deng, Zihao and Li, Wei and Liu, Yiding and You, Bo and Li, Bo and Ma, Zejun and Liu, Ziwei},
  journal={arXiv preprint arXiv:2506.20670},
  year={2025}
}

@article{zhang2025web,
  title={From Web Search towards Agentic Deep Research: Incentivizing Search with Reasoning Agents},
  author={Zhang, Weizhi and Li, Yangning and Bei, Yuanchen and Luo, Junyu and Wan, Guancheng and Yang, Liangwei and Xie, Chenxuan and Yang, Yuyao and Huang, Wei-Chieh and Miao, Chunyu and others},
  journal={arXiv preprint arXiv:2506.18959},
  year={2025}
}

@article{guo2025deepseek,
  title={Deepseek-r1: Incentivizing reasoning capability in llms via reinforcement learning},
  author={Guo, Daya and Yang, Dejian and Zhang, Haowei and Song, Junxiao and Zhang, Ruoyu and Xu, Runxin and Zhu, Qihao and Ma, Shirong and Wang, Peiyi and Bi, Xiao and others},
  journal={arXiv preprint arXiv:2501.12948},
  year={2025}
}

@misc{qwq32b,
    title = {QwQ-32B: Embracing the Power of Reinforcement Learning},
    url = {https://qwenlm.github.io/blog/qwq-32b/},
    author = {Qwen Team},
    month = {March},
    year = {2025}
}

@article{zeng2025glm,
  title={Glm-4.5: Agentic, reasoning, and coding (arc) foundation models},
  author={Zeng, Aohan and Lv, Xin and Zheng, Qinkai and Hou, Zhenyu and Chen, Bin and Xie, Chengxing and Wang, Cunxiang and Yin, Da and Zeng, Hao and Zhang, Jiajie and others},
  journal={arXiv preprint arXiv:2508.06471},
  year={2025}
}

@software{Ip_deepeval_2025,
author = {Ip, Jeffrey and Vongthongsri, Kritin},
license = {Apache-2.0},
month = aug,
title = {{deepeval}},
url = {https://github.com/confident-ai/deepeval},
version = {3.6.2},
year = {2025}
}

@article{zhang2023evaluating,
  title={Evaluating the performance of large language models on gaokao benchmark},
  author={Zhang, Xiaotian and Li, Chunyang and Zong, Yi and Ying, Zhengyu and He, Liang and Qiu, Xipeng},
  journal={arXiv preprint arXiv:2305.12474},
  year={2023}
}

%%
%% If your work has an appendix, this is the place to put it.
% \appendix

\appendix

\section{Industry Distribution}
Table~\ref{tab:ind_dist} reports the cross-market industry composition. The largest sectors are Communications (12), Consumer Staples (10), Energy (10), and Industrials (10). The remaining sectors are Consumer Discretionary (7), Health Care (6), Real Estate (3), Utilities (3), Technology (2), and Materials (1). Entries denote the number of companies in each industry–market cell; row totals are sector sizes and column totals sum to eight companies per market.

\begin{table}[h]
\centering
\caption{The company distribution with varying industries}
\setlength{\abovecaptionskip}{-0.6pt}
\setlength{\belowcaptionskip}{-0.8cm}
\setlength{\tabcolsep}{0.5mm}
% \small
\begin{tabular}{lrrrrrrrr}
\toprule
\multirow{2}{*}{\bf Industry} & \includegraphics[width=0.35cm]{icons/US.png} &\includegraphics[width=0.35cm]{icons/UK.png} & \includegraphics[width=0.35cm]{icons/CN.png} & \includegraphics[width=0.35cm]{icons/HK.png} & \includegraphics[width=0.35cm]{icons/AU.png} & \includegraphics[width=0.35cm]{icons/SG.png} & \includegraphics[width=0.35cm]{icons/MY.png} &
\includegraphics[width=0.35cm]{icons/ID.png}  \\
  & \bf US & \bf UK & \bf CN & \bf HK & \bf AU & \bf SG & \bf MY & \bf ID \\
\hline
Communications & 0 & 2 & 2 & 2 & 2 & 0 & 2 & 2 \\
Consumer Discretionary & 3 & 0 & 4 & 0 & 0 & 0 & 0 & 0 \\
Consumer Staples & 0 & 1 & 1 & 0 & 1 & 2 & 2 & 3 \\
Energy & 2 & 3 & 0 & 3 & 0 & 0 & 0 & 2 \\
Health Care & 2 & 0 & 0 & 0 & 2 & 2 & 0 & 0 \\
Industrials & 0 & 2 & 1 & 2 & 0 & 3 & 2 & 0 \\
Materials & 0 & 0 & 0 & 0 & 1 & 0 & 0 & 0 \\
Real Estate & 0 & 0 & 0 & 1 & 0 & 1 & 0 & 1 \\
Technology & 1 & 0 & 0 & 0 & 1 & 0 & 0 & 0 \\
Utilities & 0 & 0 & 0 & 0 & 1 & 0 & 2 & 0 \\

\bottomrule
\end{tabular}
\label{tab:ind_dist}
\end{table}

% A standard left-aligned paragraph column
\newcolumntype{L}{>{\raggedright\arraybackslash}p{0.24\linewidth}}
% A teletype (monospaced) left-aligned paragraph column
\newcolumntype{I}{>{\ttfamily\raggedright\arraybackslash}p{0.24\linewidth}}
% A teletype (monospaced) left-aligned tabularx column
\newcolumntype{S}{>{\ttfamily\raggedright\arraybackslash}X}

\begin{table*}[h]
\centering
\caption{Correspondence between benchmark aliases, API identifiers, and API settings}
\setlength{\abovecaptionskip}{-0.6pt}
\setlength{\belowcaptionskip}{-0.8cm}
\setlength{\tabcolsep}{0.4mm}
\small
\begin{tabularx}{\linewidth}{L I @{\hspace{12mm}}S}
\toprule
\bfseries Benchmark Alias & \bfseries API Identifier & \bfseries API Setting \\
\midrule
\rowcolor{lightgray} \multicolumn{3}{c}{\textcolor{black}{\emph{\textbf{LLM (Thinking)}}}} \\
Gemini 2.5 Pro (T)            & gemini-2.5-pro-preview-05-06 & \seqsplit{thinking\_budget=-1} \\
Deepseek-v3.2 (T)             & deepseek-v3.2-exp & \seqsplit{reasoning.enabled=True} \\
Claude-Sonnet-4.5 (T)         & claude-sonnet-4-5-20250929 & \seqsplit{thinking.type=enabled, thinking.budget\_tokens=10000} \\
Grok 4 (T)                    & grok-4-0709 & all defaults \\
OpenAI GPT-5 (T)              & gpt-5-2025-08-07 & \seqsplit{reasoning.effort=high} \\

\rowcolor{lightgray} \multicolumn{3}{c}{\textcolor{black}{\emph{\textbf{LLM (Thinking + Search)}}}} \\
Gemini 2.5 Pro (T+S)          & gemini-2.5-pro-preview-05-06 & \seqsplit{thinking\_budget=-1, tools=[google\_search]} \\
Deepseek-v3.2 (T+S)           & deepseek-v3.2-exp & \seqsplit{reasoning.enabled=True, plugins=[exa(max\_results=8)]} \\
Claude-Sonnet-4.5 (T+S)       & claude-sonnet-4-5-20250929 & \seqsplit{thinking.type=enabled, thinking.budget\_tokens=10000, tools=[web\_search\_20250305]} \\
Grok 4 (T+S)                  & grok-4-0709 & \seqsplit{search\_parameters.mode=on} \\
OpenAI GPT-5 (T+S)            & gpt-5-2025-08-07 & \seqsplit{reasoning.effort=medium, tools=[web\_search]} \\
\rowcolor{lightgray} \multicolumn{3}{c}{\textcolor{black}{\emph{\textbf{Deep Research}}}} \\
Perplexity Sonar Deep Research & sonar-deep-research & \seqsplit{reasoning.effort=high} \\
Tongyi Deep Research          & tongyi-deepresearch-30b-a3b & \seqsplit{temperature=0.6, top\_p=0.95, presence\_penalty=1.1} \\
OpenAI o3-deep-research       & o3-deep-research-2025-06-26 & \seqsplit{tools=[web\_search\_preview, code\_interpreter]} \\
\bottomrule
\end{tabularx}
\label{tab:model_ids_settings}
\end{table*}

\section{Implementation Details}

Table~\ref{tab:model_ids_settings} lists the method configurations evaluated in our benchmark. Any settings not listed were left at their default values for the corresponding method. No public API is available for Mistral Deep Research, Gemini 2.5 Pro Deep Research, or Grok 4 Deep Research; for these methods, we collected results via their official web interfaces between September~29 and October~3,~2025.

\section{Hierarchical Structure}

This section documents the hierarchical design of our markdown-based research specification. To standardize the output, we create a comprehensive template, the full structure of which is shown in Figure~\ref{fig:hstructure}. To guide generative models in populating this template, we develop a primary prompt, shown in Figure~\ref{fig:global-prompt-template}, which governs the overall research workflow and output constraints for all six sections.

This main prompt integrates detailed specifications for each section. For example, Figure~\ref{fig:section1} details the specific schema and rules for Section 1 (Company Overview). Together, this structured template and detailed prompting strategy ensure reproducibility, comparability across periods, and strict conformance to the required hierarchical output. The prompt is also adaptable; for models lacking native search capabilities (e.g., the “Thinking” method), we make minor modifications to accommodate their behavior.

\clearpage

\lstdefinestyle{markdown}{
    basicstyle=\ttfamily,
    breaklines=true,
    breakatwhitespace=true,
    frame=single,
    framesep=5pt,
    xleftmargin=10pt,
    xrightmargin=10pt,
    backgroundcolor=\color{gray!10},
    showstringspaces=false,
    tabsize=2,
    postbreak=\mbox{},
    breakindent=0pt,
    aboveskip=2em,
    belowskip=0em
}

\begin{lstlisting}[style=markdown]
## Section 1: Company Overview
### S1.1: Basic Information
| Field | Value |
| :-- | :-- |
| Company Name | |
| Establishment Date | |
| Headquarters Location | |
### S1.2: Core Competencies
| Perspective | {FY} | {FY_1} |
| :-- | :-- | :-- |
| Innovation/Product Advantages | | |
| Brand Recognition | | |
| Reputation Ratings | | |
### S1.3: Mission & Vision
| Field | Value |
| :-- | :-- |
| Mission/Vision Statement | |
| Core Values | |

## Section 2: Financial Performance
### S2.1: Income Statement
| Field | {FY} | {FY_1} | {FY_2} | Multiplier | Currency |
| :-- | :-- | :-- | :-- | :-- | :-- |
| Revenue | | | | | |
| Cost of Goods Sold | | | | | |
| Gross Profit | | | | | |
| Operating Expenses/Income | | | | | |
| Net Profit | | | | | |
| Income before income taxes | | | | | |
| Income tax expense (benefit)| | | | | |
| Interest Expense | | | | | |
### S2.2: Balance Sheet
| Field | {FY} | {FY_1} | {FY_2} | Multiplier | Currency |
| :-- | :-- | :-- | :-- | :-- | :-- |
| Total/Current/Non-Current Assets | | | | | |
| Total/Current/Non-Current Liabilities | | | | | |
| Shareholders' Equity | | | | | |
| Retained Earnings | | | | | |
| Total Equity and Liabilities | | | | | |
| Inventories | | | | | |
| Prepaid Expenses | | | | | |
### S2.3: Cash Flow Statement
| Field | {FY} | {FY_1} | {FY_2} | Multiplier | Currency |
| :-- | :-- | :-- | :-- | :-- | :-- |
| Net Cash Flow from Operations/Investing/Financing | | | | | |
| Net Increase/Decrease in Cash | | | | | |
| Dividends | | | | | |
### S2.4: Key Financial Metrics
| Field | {FY} | {FY_1} | {FY_2} |
| :-- | :-- | :-- | :-- |
| Gross/Operating/Net Profit Margin | | | |
| Current/Quick Ratio | | | |
| Debt-to-Equity | | | |
| Interest Coverage | | | |
| Asset Turnover | | | |
| Return on Equity/Assets | | | |
| Effective Tax Rate | | | |
| Dividend Payout Ratio | | | |
### S2.5: Operating Performance
| Field | {FY} | {FY_1} | {FY_2} |
| :-- | :-- | :-- | :-- |
| Revenue by Product/Service | | | |
| Revenue by Geographic Region | | | |

## Section 3: Business Analysis
### S3.1: Profitability Analysis
| Perspective | Answer |
| :-- | :-- |
| Revenue & Direct-Cost Dynamics | |
| Operating Efficiency | |
| External & One-Off Impact | |
### S3.2: Financial Performance Summary
| Perspective | {FY} | {FY_1} |
| :-- | :-- | :-- |
| Comprehensive Financial Health | | |
| Profitability and Earnings Quality | | |
| Operational Efficiency | | |
| Risk Identification and Early Warning| | |
| Future Financial Performance Projection | | |
### S3.3: Business Competitiveness
| Perspective | {FY} | {FY_1} |
| :-- | :-- | :-- |
| Business Model | | |
| Market Position | | |

## Section 4: Risk Factors
### S4.1: Risk Factors
| Perspective | {FY} | {FY_1} |
| :-- | :-- | :-- |
| Market/Operational/Financial/Compliance Risks| | |

## Section 5: Corporate Governance
### S5.1: Board Composition
| Name | Position | Total Income |
| :-- | :-- | :-- |
| | | |
### S5.2: Internal Controls
| Perspective | {FY} | {FY_1} |
| :-- | :-- | :-- |
| Risk Assessment Procedures | | |
| Control Activities | | |
| Monitoring Mechanisms | | |
| Identified Material Weaknesses/Deficiencies | | |
| Effectiveness | | |

## Section 6: Market Performance
### S6.1: Stock Performance
| Field | {CY} | {CY_1} |
| :-- | ----:| ----:|
| Lowest/Highest Adjusted Closing Price | | |
| Total Log Return | | |
| Log Excess Return | | |
| Maximum Drawdown | | |
| Annualized Volatility | | |
### S6.2: News Sentiment Analysis
| Field | {CY} | {CY_1} |
| :-- | :-- | :-- |
| Top 1/2/3 Positive Window Date/Summary | | |
| Top 1/2/3 Negative Window Date/Summary | | |
### S6.3: Market Reaction to News
| Field | {CY} | {CY_1} |
| :-- | :-- | :-- |
| Top 1/2/3 Positive Window Date/CAR/Summary | | |
| Top 1/2/3 Negative Window Date/CAR/Summary | | |
### S6.4: Price-to-Earnings (P/E) Ratio
| Field | Value as of {DATE} |
| :-- | :-- |
| Adjusted Closing Price | |
| Diluted EPS & P/E Ratio | |
\end{lstlisting}
\captionof{figure}{Complete hierarchical structure for 6 main sections, 18 subsections and 18 markdown tables}
\label{fig:hstructure}

\begin{lstlisting}[style=markdown]
Search and analyze the annual reports({LANGUAGE} version) from FY{FY_1}{FY_A} and FY{FY} for {COMPANY} listed on {STOCK_MARKET}. Assuming today's date is 2025-09-20, generate a research report on this company. The report should be in markdown format and must follow the given structure below and include all the tables below.

As defined in the `Research Scope` of each section, annual reports are sufficient for conducting analyses from Section 1 to Section 5. However, to complete Section 6, you should diligently search for additional information such as news, stock prices, or any relevant data that can support your research. Further requirements can be derived from the `Research Scope` of each section.

`Research Language`: Please make sure your answer is written in {LANGUAGE}. Note that always leave the section headers and table headers in English.

`Research Output Format`: Your response must consist ONLY of the section and subsection headers and the completed markdown tables as defined below. No text outside of tables is permitted. All analyses and summaries must be written within table cells, using detailed content (typically 4-8 sentences or structured bullet points).


## Section 1: Company Overview

*[Details and table structure omitted for brevity]*

## Section 2: Financial Performance

*[Details and table structure omitted for brevity]*

## Section 3: Business Analysis

*[Details and table structure omitted for brevity]*

## Section 4: Risk Factors

*[Details and table structure omitted for brevity]*

## Section 5: Corporate Governance

*[Details and table structure omitted for brevity]*

## Section 6: Market Performance

*[Details and table structure omitted for brevity]*
\end{lstlisting}
\captionof{figure}{Research Task Prompt for Sections 1–6}
\label{fig:global-prompt-template}

\begin{lstlisting}[style=markdown]
## Section 1: Company Overview
This section provides a concise overview of the company, including its basic information, industry background, key strengths, and strategic direction.

`Research Scope`: Focus on the FY{FY} annual report for FY{FY} data, and the FY{FY_1} annual report for FY{FY_1} data for {COMPANY} listed on {STOCK_MARKET}.

### S1.1: Basic Information
This subsection provides fundamental information about the company's identity.

Create a table with Markdown format with Field and Value headers for the following items:

1. Company Name
2. Establishment Date
3. Headquarters Location (City and Country)

| Field | Value |
| :---- | :---- |
| Company Name |  |
| Establishment Date |  |
| Headquarters Location |  |

### S1.2: Core Competencies
This section provides information about the company's core competencies. Create a summary in the table below for each perspective, offering readers insight into the company's competitive strengths and unique value propositions.

| Perspective | {FY}  | {FY_1} |
| :---- | :---- | :---- |
| Innovation Advantages |  |  |
| Product Advantages |  |  |
| Brand Recognition |  |  |
| Reputation Ratings |  |  |

### S1.3: Mission & Vision
This section provides information about the company's purpose and long-term goals. Create a summary in the table below for each perspective in the single cell, offering readers a clear understanding of the company's strategic direction and aspirations.

| Field | Value |
| :---- | :---- |
| Mission Statement |  |
| Vision Statement |  |
| Core Values |  |
\end{lstlisting}
\captionof{figure}{Section 1 specification: Company Overview—scope, subsections, and table schemas}
\label{fig:section1}

\end{document}